 \documentclass[10pt,letterpaper,compsoc,conference]{iiswc26}

\usepackage{cite}
\usepackage{amsmath,amssymb,amsfonts}
\usepackage{algorithmic}
\usepackage{graphicx}
\usepackage[dvipsnames]{xcolor}
\usepackage[final]{microtype}
\usepackage[italic]{mathastext}
\usepackage{libertine}
\usepackage[T1]{fontenc}
\usepackage{textcomp}
\usepackage[varqu,varl]{zi4}
\usepackage[all]{nowidow}
\usepackage[auth-lg,affil-it]{authblk}
\usepackage[keeplastbox]{flushend}
\usepackage{fancyhdr}

\usepackage{xcolor}
\usepackage{url}
\usepackage{balance}

\newcommand{\todo}[1]{}
\newcommand{\add}[1]{}

\usepackage{xcolor}

\definecolor{takeawaygray}{RGB}{225,225,225}
\definecolor{reccoral}{RGB}{246,221,213}
\definecolor{stageblue}{HTML}{434abc}
\newcommand{\takeaway}[2]{%
  \noindent\colorbox{takeawaygray}{\textbf{Insight #1:}}~\textit{#2}\par\vspace{0.4em}
}

\newcommand{\recommendation}[2]{%
  \noindent\colorbox{reccoral}{\textbf{Recommendation #1:}}~\textit{#2}\par\vspace{0.4em}
}

\usepackage{enumitem}
\usepackage{booktabs}
\usepackage{multirow}
\usepackage{graphicx}
\usepackage{xcolor}
\usepackage{pifont}
\usepackage{tikz}
\usepackage{enumitem}
\usepackage{float}
\usepackage{svg}
\newcommand{\cmark}{\textcolor{green!60!black}{\checkmark}}
\newcommand{\xmark}{\textcolor{red!75!black}{\ding{55}}}

\newcommand*\circled[1]{%
  \tikz[baseline=(char.base)]{
    \node[
      shape=circle,
      draw,
      line width=0.9pt,
      inner sep=1.8pt
    ] (char) {\small\bfseries #1};
  }%
}
\usepackage{caption} 

\usepackage{fancyhdr}

\begin{document}


\title{Agent Memory: Characterization and System Implications of Stateful Long-Horizon Workloads}

\renewcommand\Authsep{\qquad}
\renewcommand\Authand{\qquad}
\renewcommand\Authands{\qquad}





\author[ ]{%
  Yasmine Omri$^{1\star}$ \quad
  Ziyu Gan$^{2\star}$ \quad
  Zachary Broveak$^{1}$ \\
  Robin Geens$^{3}$ \quad
  Zexue He$^{1}$ \quad
  Alex Pentland$^{1,4}$ \quad
  Marian Verhelst$^{3}$ \\
  Tsachy Weissman$^{1}$ \quad
  Thierry Tambe$^{1}$\\
  {$^\star$\textit{Equal contribution}}
}

\affil[ ]{$^{1}$Stanford University \qquad $^{2}$Independent Researcher}
\affil[ ]{$^{3}$MICAS, KU Leuven \qquad $^{4}$Massachusetts Institute of Technology}

\twocolumn[{%

  \renewcommand\twocolumn[1][]{#1}%

  \maketitle
  \pagestyle{empty}
  \thispagestyle{empty}

  \pagestyle{plain}

  \vspace{-1.25em}

  \begin{center}
    \hspace{3em}\includegraphics[width=1.05\textwidth]{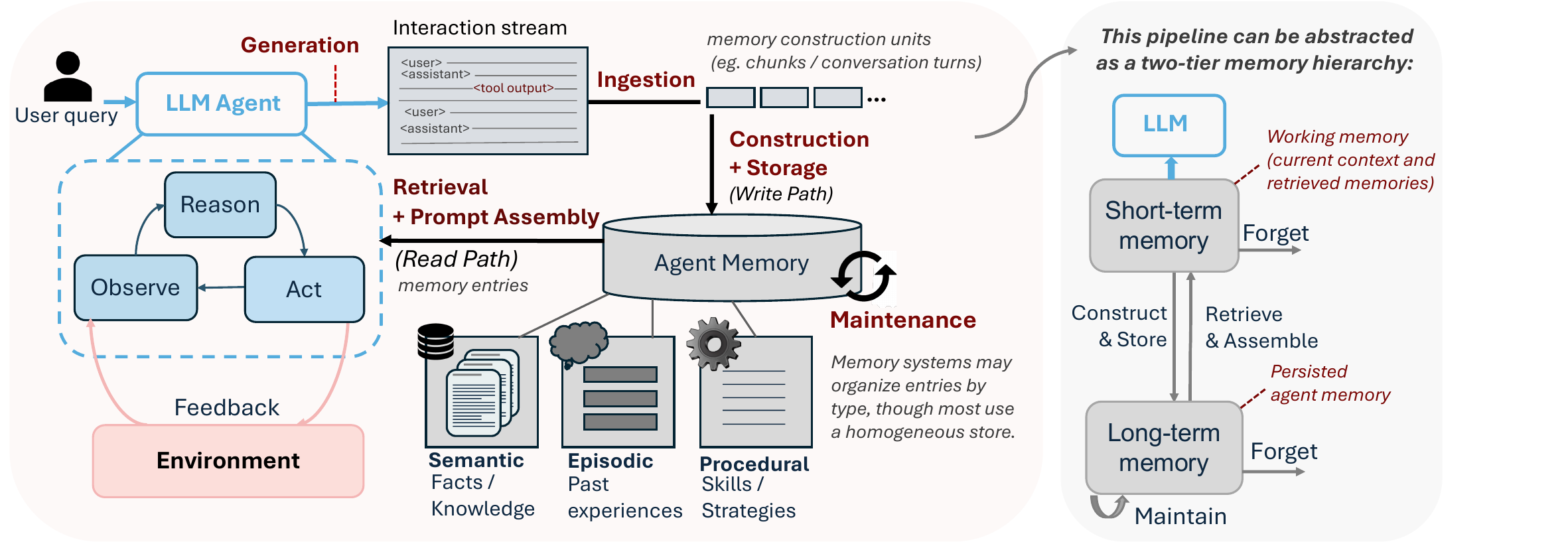}

    \vspace{-0.25em}

    \captionsetup{type=figure}
    \caption{\textbf{Agent memory}
    gives rise to short-term working memory and long-term memory:
    the agent retrieves relevant long-term state into its active context, updates memory
    after interaction, and maintains stored knowledge
    over time.}
    \label{fig:overview}
  \end{center}
}]





\begin{abstract}

\renewcommand{\thefootnote}{\relax}
\footnotetext{\hspace{-1.8em}\textit{Correspondence to: \texttt{yomri@stanford.edu}, \texttt{ziyu.gan@outlook.com}}}
\renewcommand{\thefootnote}{\arabic{footnote}}

LLM agents are increasingly deployed on long-horizon tasks requiring sustained reasoning over extended interaction histories. Realizing this at scale requires agents to persistently store, retrieve, and update their own memory across sessions. A rich ecosystem of agent memory systems has emerged spanning flat retrieval, LLM-mediated extraction, consolidating fact stores, and agentic control flows. Yet, their system-level behavior remains uncharacterized. We present the first systems characterization of agent memory. First, we introduce a system-oriented taxonomy classifying agent memory systems along four axes. Second, we build a phase-aware profiling harness attributing cost to construction, retrieval, and generation. Third, we characterize ten representative systems across two benchmark suites, uncovering how design choices shift cost across the write and read paths. Finally, we derive 10 system recommendations covering construction scheduling, capability floors, amortization via query volume, freshness-latency tradeoffs, and fleet-scale management.

\end{abstract}

\section{Introduction}

Large language models (LLMs) are increasingly deployed as autonomous agents capable of
multi-step reasoning, tool use, and persistent interaction over extended time horizons.
Stateful execution enables LLMs to act as personal assistants that track user preferences across weeks of
dialogue, code agents that maintain project context across editing sessions, deep-research agents that synthesize evidence across long investigation
trajectories, and
enterprise systems that integrate structured business data with ongoing conversational
history~\cite{park2023generative,wang2023agentsurvey,packer2023memgpt}. In these regimes, LLM agents accumulate state that far exceeds what a single inference
context can hold. As such, realizing these long-horizon capabilities at scale requires agents to persistently store, selectively
retrieve, and incrementally update their own memory across sessions, as illustrated by Fig.~\ref{fig:overview}.

\textbf{Agent memory generalizes RAG from static retrieval to mutable state management.} The evolution toward stateful agents reflects a progression in how LLMs manage knowledge.
Early systems treated knowledge as purely parametric, encoded in model weights at training
time. Retrieval-augmented generation (RAG) decoupled knowledge from parameters by
attaching LLMs to static external corpora, enabling selective access to non-parametric
state at inference time ~\cite{lewis2020rag,karpukhin2020dpr}.
Long-horizon agents extend this further: the memory corpus is no longer a fixed document
collection that is preprocessed and indexed a priori, but a mutable state produced by the agent’s own interaction stream, often per user, and may be appended, summarized, consolidated, linked, or rewritten across sessions as new evidence arrives. This shift transforms memory from a passive retrieval target
into an active systems component with a write path, a search path, and an ongoing maintenance policy.


A natural baseline is to preserve the full interaction history in the model's context window, relying on its native long-context capability as the memory mechanism. While context limits have rapidly scaled from tens of thousands to over one million tokens in just the last two years ~\cite{team2024gemini}, this approach faces three fundamental limits. Multi-session interactions and tool traces inevitably exceed any fixed context budget. Prefill cost grows quadratically with history length, and prefix caching fails across sessions due to eviction and KV-cache memory pressure. Reasoning and recall fidelity degrade significantly in long sequences. Models exhibit a ``U-shaped'' performance curve where facts in the middle of the context are routinely lost~\cite{liu2023lostmiddle}, and their \textit{effective} context length for complex reasoning is often a fraction of the claimed window~\cite{hsieh2024ruler}. External memory systems resolve these constraints by persisting state externally and retrieving only a query-relevant subset, bounding per-query prefill and enabling stateful execution for long-horizon agents.

The last year has seen a surge of agent memory systems capable of ingesting, encoding, and persisting an interaction stream into a queryable state. Their designs differ dramatically in construction mechanism, database structure, retrieval flow, and update policy~\cite{packer2023memgpt,chhikara2025mem0,gutierrez2025fromragtomemory,edge2025graphrag,xu2025amem,wang2025mirix}. Existing benchmarks evaluate these systems purely on downstream accuracy~\cite{maharana2024locomo,wu2024longmemeval,hu2025memoryagentbench,wei2025evomemory}, leaving their system-level behavior uncharacterized. Yet, these design choices drastically and asymmetrically redistribute cost: a system that compresses history into atomic facts slashes query-time prompt length but pays orders-of-magnitude more in construction-time LLM prefill; a graph-based memory improves relational retrieval while multiplying embedding traffic and storage footprint. These costs are invisible to accuracy metrics, yet critically dominate at deployment scale. In this paper, we present the first systems study of agent memory algorithms, spanning the full design space. Our study is organized around three research questions: 
\begin{enumerate}[label=\arabic*)]
  \item \textbf{\textit{What are the key
tradeoffs among emerging paradigms for constructing and using memory in long-horizon agent workloads?}}
  \item \textbf{\textit{What computational costs and system opportunities do these
workloads expose, and what demands do they place on deployment infrastructure?}}
  \item \textbf{\textit{How do specific algorithmic design choices concerning the construction mechanism, storage organization, retrieval policy, and update strategy shape observable system
characteristics such as utilization, bandwidth, latency, and scalability?}}
\end{enumerate}

Our analysis reveals that agent memory workloads impose system needs beyond traditional LLM serving, making the following four contributions:
\begin{itemize}
\item \textbf{A system-oriented taxonomy} classifying ten agent memory systems along four axes (construction, storage, retrieval, and mutability) with predicted cost signatures per paradigm.
\item \textbf{A phase-aware profiling harness (open-sourced)} tracking tokens, model calls, GPU utilization, and latency across construction, retrieval, and generation.
\item \textbf{A comprehensive systems characterization} spanning construction cost shape, capability thresholds, amortization structure, freshness scheduling, footprint growth, and retrieval tail behavior.
\item \textbf{Ten system recommendations} for agent memory serving infrastructure, scheduling, and system selection.
\end{itemize}

\section{Agent Memory Paradigms}
\label{sec:Paradigms}

This section introduces the execution structure and design space of agent memory
systems. We first decompose agent memory execution into its core runtime stages
(Sec.~\ref{sec:memory-pipeline}), then categorize recent agent memory systems into
four paradigms according to how they construct, organize, and retrieve persistent
state (Sec.~\ref{sec:memory-taxonomy}). Table~\ref{tab:agent memory-taxonomy}
summarizes the representative systems considered in this study.

\subsection{Agent Memory Execution Pipeline}
\label{sec:memory-pipeline}

Agent memory systems  transform an interaction stream into persistent state that can
be reused across future agent invocations. Although implementations differ widely,
most systems can be decomposed into seven stages: ingestion, memory construction,
storage, retrieval, prompt assembly, generation, and maintenance.
\par\textcolor{stageblue}{\textbf{Ingestion.}}
The ingestion stage receives the raw interaction stream, which may include user-assistant dialogue, tool calls, documents, execution traces, and environmental feedback. Ingestion defines the unit of memory processing: individual turns, fixed-size chunks, sliding windows, or complete sessions. This choice determines write granularity and the number of downstream construction operations.
\par\textcolor{stageblue}{\textbf{Memory construction.}}
We use \emph{memory construction} to denote the write-path transformation from raw interaction history into persistent memory records. Construction takes one of four forms: \textbf{absent} (history retained directly in the prompt), \textbf{deterministic} (chunking or indexing without any LLM), \textbf{LLM-mediated} (extraction of facts, summaries, or structured records at predefined points, optionally with add, update, or delete consolidation over prior memories), or \textbf{agentic} (LLM-controlled writes, tool invocation, and record mutation)~\cite{packer2023memgpt,chhikara2025mem0,xu2025amem,wang2025mirix}.
\par\textcolor{stageblue}{\textbf{Storage.}}
The storage stage persists constructed memories in a system-specific agent database, which is not necessarily a vector store. Memories may be stored as raw prompt buffers, inverted indices, dense vector indices, knowledge graphs, or multi-store combinations coupling lexical, vector, graph, and structured metadata. The storage substrate determines memory footprint, update cost, index rebuild requirements, and available retrieval operators.
\par\textcolor{stageblue}{\textbf{Retrieval.}}
The retrieval stage selects memory records relevant to the current query. Retrieval ranges from a single-shot sparse or dense search to multi-stage pipelines combining sparse search, dense search, graph traversal, reranking, and filtering. In agentic systems, retrieval becomes a tool-mediated control loop where the LLM decides when to search, which memory type to query, and whether retrieved evidence is sufficient.
\par\textcolor{stageblue}{\textbf{Prompt assembly.}}
Retrieved memories are serialized into the final model context, yielding the prefill length of the final answer-generation call.
\par\textcolor{stageblue}{\textbf{Generation.}}
The generation stage invokes the task LLM on the current query and assembled memory context. All agent memory systems converge here, but arrive with different prompt lengths and prior orchestration costs. Two systems may produce similar answers while imposing very different write-path and read-path workloads.
\par\textcolor{stageblue}{\textbf{Maintenance.}}
The maintenance stage updates memory over the agent's lifetime through deduplication, consolidation, conflict resolution, forgetting, compression, or re-embedding. In many current systems, maintenance is weak or absent: memory accumulates indefinitely with no freshness or pruning policy, which becomes critical for long-lived agents where state grows continuously and may become redundant or stale.

\subsection{Taxonomy of Agent Memory Paradigms}
\label{sec:memory-taxonomy}

Based on how the agent builds and interacts with the memory, we classify agent memory systems into four paradigms: \emph{long-context memory},
\emph{flat RAG memory}, \emph{structure-augmented RAG memory}, and
\emph{agentic control flow}. Table~\ref{tab:agent memory-taxonomy} summarizes
representative systems across these paradigms and records their construction
pipeline, agent database, retrieval pipeline, and mutability. For our workload characterization, we select ten agent memory systems spanning all four paradigms and their subdivisions. The remainder of this section describes each paradigm and its representative agent memory systems in turn.

\begin{table*}[t]
\centering
\caption{\textbf{Agent Memory Systems.}
Categorization of representative agent memory systems into four paradigms with their
construction pipeline, agent database, retrieval pipeline, and mutability. In the
retrieval columns, ``LLM'' indicates an LLM call before final answer generation.}
\label{tab:agent memory-taxonomy}
\resizebox{\textwidth}{!}{
\begin{tabular}{llcccclclc}
\toprule
\multirow{2}{*}{\textbf{Paradigm}} &
\multirow{2}{*}{\textbf{Agent Memory System}} &
\multicolumn{3}{c}{\textbf{Construction Pipeline}} &
\multicolumn{2}{c}{\textbf{Agent DB}} &
\multicolumn{2}{c}{\textbf{Retrieval Pipeline}} &
\multirow{2}{*}{\textbf{Mutability}} \\
\cmidrule(lr){3-5} \cmidrule(lr){6-7} \cmidrule(lr){8-9}
& & Agent Ctrl. & LLM & Embed & Struct. & Store & LLM & Flow & \\
\midrule

I: Long-context memory
& long\_context
& \xmark & \xmark & \xmark
& \xmark & raw context
& \xmark & passthrough
& append \\

\midrule

\multirow{2}{*}{II: Flat RAG memory}
& BM25
& \xmark & \xmark & \xmark
& \xmark & inverted index
& \xmark & lexical top-$k$
& append \\

& embedRAG
& \xmark & \xmark & \cmark
& \xmark & dense store
& \xmark & dense top-$k$
& append \\

\midrule

\multirow{2}{*}{\shortstack{III.a: Structure-aug. RAG\\append-only}}
& GraphRAG
& \xmark & \cmark & \cmark
& \cmark & graph store
& optional & graph expand
& append \\

& HippoRAG v2
& \xmark & \cmark & \cmark
& \cmark & multi-view graph
& optional & PPR rerank
& append \\

\midrule

\multirow{2}{*}{\shortstack{III.b: Structure-aug. RAG\\consolidating}}
& Mem0
& \xmark & \cmark & \cmark
& \cmark & dense store
& \xmark & fact top-$k$
& consolidate \\

& SimpleMem
& \xmark & \cmark & \cmark
& \cmark & hybrid store
& \cmark & iterative hybrid
& consolidate \\

\midrule

\multirow{3}{*}{IV: Agentic control flow}
& A-Mem
& \cmark & \cmark & \cmark
& \cmark & graph store
& \xmark & graph-structured map-reduce
& mutate \\

& Letta
& \cmark & \cmark & optional
& \cmark & blocks+archive
& \cmark & tool calls
& mutate \\

& MIRIX
& \cmark & \cmark & \cmark
& \cmark & multi-store DB
& \cmark & routed typed-memory search
& mutate \\

\bottomrule
\end{tabular}}
\end{table*}

\noindent\textbf{PARADIGM I: Long-context memory.}
Long-context memory uses the model context itself as the memory substrate. Prior
interactions are retained as raw text and passed directly, or after truncation, to
the LLM.

\begin{enumerate}[label=\protect\circled{\arabic*}, leftmargin=*, itemsep=2pt]
    \item \textbf{long\_context~\cite{liu2023lostmiddle,hsieh2024ruler}.}
    This agent memory system performs no memory construction and stores no external
    representation. The full interaction history is inserted into the prompt, making retrieval equivalent to passthrough with maximal final-prompt growth. \add{mention that we use gpt-4o-mini + 4.1mini and how gpt-4o-mini doesn't comfortably accommodate all benchmark context lengths}
\end{enumerate}

\noindent\textbf{PARADIGM II: Flat RAG memory.}
Flat RAG memory applies a deterministic indexing pipeline (a lexical or vector index) to raw interaction chunks. Construction is append-only and does not involve an LLM. Retrieval is typically single-shot. This paradigm is close to classical RAG, except the corpus is produced by the agent's own interaction stream rather than a
static document collection.

\begin{enumerate}[label=\protect\circled{\arabic*}, leftmargin=*, itemsep=2pt, resume]
    \item \textbf{BM25~\cite{robertson2009probabilistic}}
    chunks the interaction history into an inverted index. Construction uses no LLM or embedding model. Retrieval is lexical top-$k$. BM25 represents the lowest-complexity external-memory design in our suite.

    \item \textbf{embedRAG~\cite{lewis2020rag,karpukhin2020dpr,johnson2019billion}}
    chunks and embeds the interaction history into a dense index. Retrieval embeds the query and returns nearest-neighbor chunks. Both paths route through the embedding service, but the write path remains single-shot and append-only.
\end{enumerate}
\noindent\textbf{PARADIGM III: Structure-augmented RAG memory.}
These systems use an LLM as a fixed extractor to derive facts, summaries, entities, triples, or graph records from the interaction stream~\cite{sarthi2024raptor,edge2025graphrag,gutierrez2025fromragtomemory,chhikara2025mem0,rasmussen2025zep}. The LLM is called at predefined points to produce structured outputs, not as a free-form controller. We distinguish append-only variants, which enrich the retrieval substrate without revising prior records, from consolidating variants, which distill or update records over time.\\
\textit{PARADIGM III.a: Append-only structure-augmented RAG.}
\begin{enumerate}[label=\protect\circled{\arabic*}, leftmargin=*, itemsep=2pt, resume]
\item \textbf{GraphRAG~\cite{edge2025graphrag}.}
Uses an LLM to extract entity-relation structure from chunks, storing entities, relationships, and source chunks alongside dense representations. Retrieval combines dense lookup with graph expansion, exposing LLM-based extraction on the write path and graph traversal on the read path.
\item \textbf{HippoRAG v2~\cite{gutierrez2025fromragtomemory}.}
Builds an associative memory over passages, entities, and facts via OpenIE-style extraction, with each view embedded separately and linked through a graph. Retrieval propagates relevance via Personalized PageRank over a coupled passage-entity-fact graph before reranking.
\end{enumerate}
\textit{PARADIGM III.b: Consolidating structure-augmented RAG.}
\begin{enumerate}[label=\protect\circled{\arabic*}, leftmargin=*, itemsep=2pt, resume]
\item \textbf{Mem0~\cite{chhikara2025mem0}.}
Rewrites conversation turns into atomic facts and resolves each against existing memories via ADD, UPDATE, or DELETE decisions. Retrieval is compact: query embedding matched against the fact store with no retrieval-side LLM, separating a semantically expensive write path from a lightweight read path.
\item \textbf{SimpleMem~\cite{liu2026simplemem}.}
Processes dialogue through sliding windows into distilled memory entries with semantic, lexical, and structured views. Retrieval is hybrid, combining semantic search, keyword search, and structured filtering, with optional LLM planning and reflection rounds to expand evidence.
\end{enumerate}
\noindent\textbf{PARADIGM IV: Agentic control flow.}
Agentic memory systems expose memory operations to the LLM as part of its decision loop, turning construction and retrieval into variable-depth control flow~\cite{packer2023memgpt,shinn2023reflexion,xu2025amem,wang2025mirix}. Unlike Paradigm III, the LLM decides when to write, which tool to invoke, and whether retrieved evidence is sufficient.
\begin{enumerate}[label=\protect\circled{\arabic*}, leftmargin=*, itemsep=2pt, resume]
\item \textbf{A-Mem~\cite{xu2025amem}.}
Treats memory as an evolving note graph. Each new note is matched against nearby memories and an LLM evolution step updates links, metadata, and neighboring records, exposing stateful mutation on the write path.
\item \textbf{Letta~\cite{letta2024}.}
Implements the MemGPT~\cite{packer2023memgpt} abstraction, separating compact core memory from archival memory and exposing reads and writes as LLM-callable tools. Memory access is an action selected by the agent, making latency dependent on when and how the LLM invokes memory operations.

\item \textbf{MIRIX~\cite{wang2025mirix}.}
Routes incoming information to type-specific extractors across episodic, semantic, procedural, and resource memory stores. Retrieval uses type-specific operators including vector search, temporal filtering, and LLM-based synthesis, stressing both the LLM control path and the heterogeneous database layer.
\end{enumerate}

\section{Workload Suite and Profiling Harness}
\label{sec:workload-suite}

\subsection{Workload Suite}
\label{sec:workload-suite-benchmarks}

\textbf{MemoryAgentBench.}
We use MemoryAgentBench (MAB) as our primary benchmark~\cite{hu2025memoryagentbench}. MAB converts long-context tasks into incremental multi-turn streams and evaluates four memory competencies: accurate retrieval, test-time learning, long-range understanding, and selective forgetting. Table~\ref{tab:mab-suite} summarizes the task families. Each sample is fed incrementally in 4096-token chunks. We use the full suite to study quality trends, but center our systems characterization on \texttt{LongMemEval\_S\_*}, the most widely adopted long-term conversational memory setting in MAB. This workload contains five samples, each with approximately 360 K tokens of history and 60 queries (300 total), cleanly separating construction over a long interaction history from repeated query-time retrieval and generation.

Because agent memory systems expose many implementation degrees of freedom, a perfectly apples-to-apples comparison is not possible. We therefore standardize the dominant controllable factors while preserving each system's native execution model. Within each model configuration, all systems use the same generation LLM and the same embedding backbone when LLM calls and/or embeddings are required. Histories are streamed in 4096-token chunks, matching the MAB protocol, but we retain each system's native buffering, parallel ingestion, and cross-chunk consolidation behavior, since these mechanisms are part of the system being characterized. At query time, we cap retrieval at 10 memory entries for all systems. In local-model settings with restricted context windows, if a system overflows the context budget, we reduce retrieval to 5 entries. For Letta, we additionally use 512-token ingestion chunks to avoid repeated context overflow during tool-mediated memory construction. \\
Several systems were originally designed for conversational agents rather than the full diversity of MAB tasks. To avoid penalizing them for prompt-format mismatch rather than memory quality, we make minimal task-adaptation changes. For dialogue-oriented systems such as SimpleMem, we update the extraction prompts so that they can consume non-dialogue histories and preserve task-relevant facts. For A-Mem, whose native parser assumes each conversational turn forms a coherent memory note, we aggregate consecutive fine-grained parsed units on non-dialogue datasets so that memory notes are comparable in granularity to conversation-derived notes. For Mem0 on in-context-learning datasets, we replace the default personalization-oriented extraction prompt with an ICL-aware prompt that preserves integer labels verbatim, since these labels are the task signal. For Letta with local models, we cap repeated tool calls during construction and retrieval and lightly specialize the system prompt to the task format. These adaptations are restricted to interface and prompt-level compatibility changes. For Letta, the tool-call caps prevent local models from emitting hundreds of duplicate memory tool calls, and task-specific system prompts ensure that task-critical symbols, such as ICL class labels, are stored rather than summarized away. For A-Mem, local backbones use the plain-text prompt variant, while API backbones retain the original JSON-schema prompt. For Mem0, the ICL-aware extractor is applied only to ICL datasets; all other ingestion behavior is unchanged.

\begin{table}[htbp]
\centering
\caption{\textbf{MemoryAgentBench workloads.}
MAB groups memory-agent workloads into four task categories. AvgL denotes the
average input length reported by MAB.}
\label{tab:mab-suite}
\begin{tabular}{p{0.28\columnwidth}p{0.4\columnwidth}p{0.2\columnwidth}}
\toprule
\textbf{Task Category} & \textbf{Datasets} & \textbf{AvgL.} \\
\midrule
Accurate retrieval
& SH-Doc QA, MH-Doc QA, \texttt{LongMemEval\_S\_*}, EventQA
& 197K--534K \\
\midrule
Test-time learning
& BANKING77, CLINC150, NLU, TREC, MovieRec
& 103K--1.44M \\
\midrule
Long-range understanding
& $\infty$Bench-Sum, Detective QA
& 124K--172K \\
\midrule
Selective forgetting
& FactConsolidation-SH/MH
& 262K \\
\bottomrule
\end{tabular}
\end{table}

\textbf{MemoryArena.}
We additionally use MemoryArena~\cite{he2026memoryarena}, which contains multi-session tasks, where later sessions depend on information established earlier. We use its session structure to surface a systems property static benchmarks hide: freshness-latency tradeoff in Sec.~\ref{sec:freshness}. When construction is slower than session arrival, the system must either block queries until writes commit or serve stale memory, exposing scheduling and write-prioritization opportunities.

\subsection{Serving and Hardware Setup}
\label{sec:serving-setup}

We evaluate two construction regimes. In the \emph{remote-construction} regime,
construction uses OpenAI-hosted models (GPT-4o-mini or GPT-4.1-mini for LLM calls,
text-embedding-3-small for embeddings), matching most systems' original
release configurations. In the \emph{local-construction} regime, all models are
served locally on one NVIDIA H100 80\,GB HBM3 GPU via vLLM~\cite{kwon2023efficient},
enabling measurement of hardware utilization, energy, and phase-level bottlenecks.
Each system runs in an isolated SLURM job, with one GPU and six Intel Xeon Platinum 8480C cores. The LLM ladder comprises Qwen3-32B,
Qwen3-14B, Qwen3-8B, and Qwen3-1.7B; the embedding model is Qwen3-Embedding-0.6B.
Qwen3-32B and Qwen3-14B use FP8 weight quantization to fit within a single 80\,GB
serving stack with sufficient KV-cache headroom for long-prefill
execution~\cite{micikevicius2022fp8,vllmfp8}. Smaller models use BF16. The LLM
server is allocated 75\% of GPU memory and the embedding server 15\%. Thinking mode
is disabled for all calls to avoid context overflow from verbose reasoning traces. 
To validate that our findings extend beyond single-device serving,
we additionally profile selected systems under tensor-parallel serving with
Qwen3-235B (TP=8) on 8$\times$H100 GPUs (Sec.~\ref{sec:construction-shape}).
\subsection{Profiling Harness}
\label{sec:profiling-harness}

We build a phase-aware profiling harness (open-sourced) that attributes cost to three logical phases: memory construction, retrieval, and generation. The harness records API telemetry and hardware telemetry on the same monotonic timeline, enabling per-phase attribution of token volume, call structure, latency, utilization, and energy. 

\textbf{API telemetry.}
We instrument every chat-completion and embedding request (whether remote or local through vLLM) issued by an agent memory system, recording call type, source label, start/end time, latency, prompt tokens, completion tokens, embedding input tokens, and number of embedded sequences. Calls are tagged with the active phase and the chunk, window, turn, or query index that triggered them. 

\textbf{Hardware telemetry.}
A polling monitor runs a separate thread outside the measured workflow, sampling GPU counters via NVML and DCGM-class GPM counters without injecting kernels, recording device power, GPU utilization, VRAM footprint, SM activity, tensor-core activity, and HBM bandwidth. Samples are aligned with phase markers, and phase energy is computed by integrating device power over each interval. Together, the API and hardware telemetry characterize each agent memory system as a workload: call volume, prompt and embedding traffic, GPU energy, and how control flow manifests as construction, retrieval, and user-visible generation latency.
\section{Characterizing Agent Memory Workloads}
\label{sec:characterization}
This section characterizes agent memory as a systems workload, tracing how design choices shift cost across construction, retrieval, and generation, and how operators should select and host agent memory systems for SLO-bound serving and fleet-scale deployment.

\subsection{Why Agent Memory}
\label{sec:why-memory}

\begin{figure}[t]
    \centering
     \includegraphics[width=1\columnwidth]{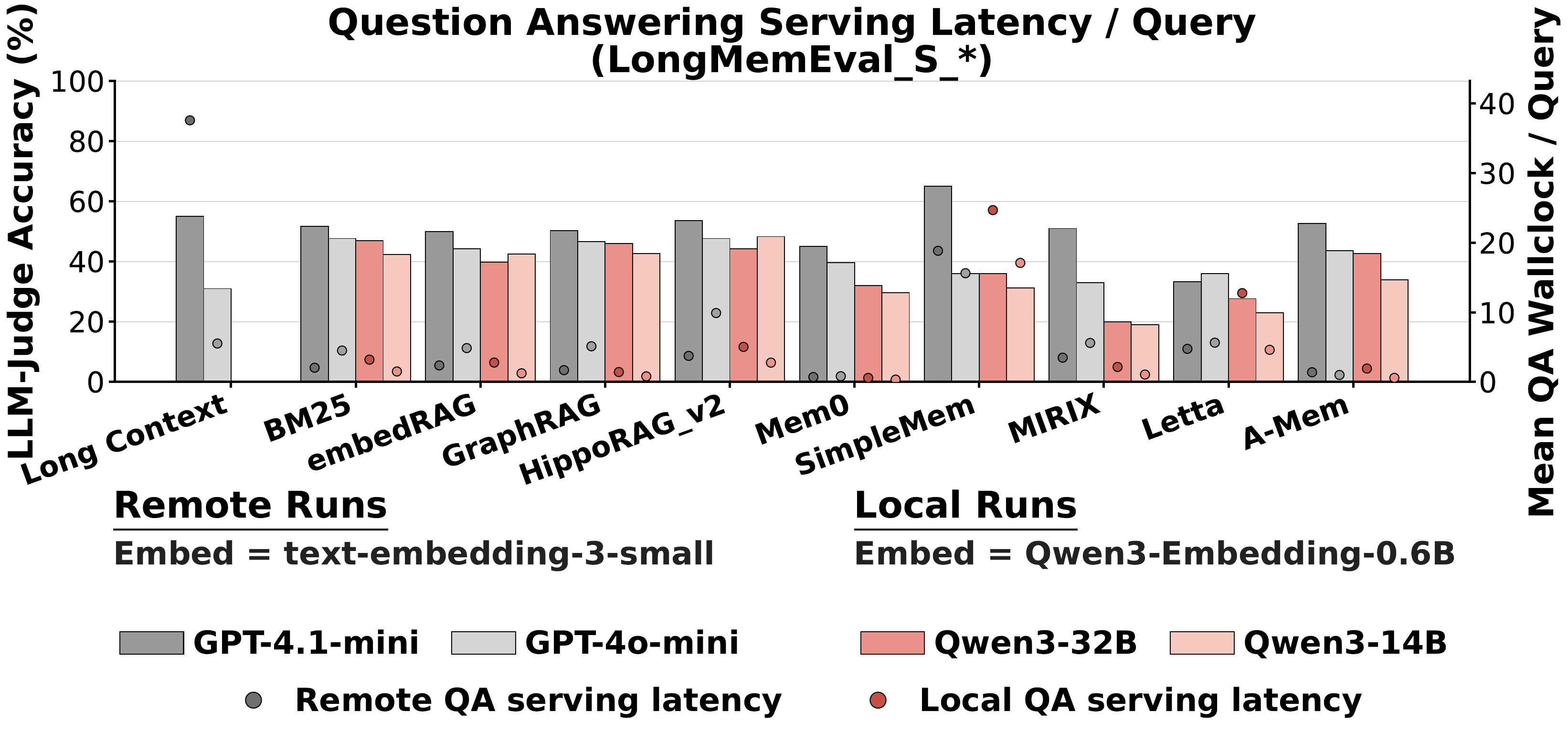}
    \caption{\textbf{Long-context prompting vs. external agent memory.}
    Per-query serving latency (retrieval + generation, excluding construction) on
    \texttt{LongMemEval\_S\_*}. Remote construction via OpenAI API, local construction via vLLM.}
    \label{fig:serving}
\end{figure}

To quantify the benefits of long-horizon agent memory systems over full-history baselines, we measure the task accuracy and execution times of running \texttt{LongMemEval\_S\_*} for a multitude of agent memory systems.
Fig.~\ref{fig:serving} plots per-query serving latency against accuracy, where each system ingests 5 samples of $\sim$360\,K tokens of
history each (300 total queries). Construction cost is excluded here
and addressed in Sec.~\ref{sec:hidden-construction-cost}. Most agent memory systems
serve queries in a fraction of long-context wall time while matching or exceeding
its accuracy: Mem0 achieves $<$0.1\,s per query versus $\sim$38\,s for long-context
(GPT-4.1-mini, OpenAI API, averaged over 300 queries). The taxonomy predicts this
spread directly. Paradigm~I pays full-history prefill at every query. Paradigm~II
eliminates the LLM from construction and performs deterministic top-$k$ retrieval,
yielding the lowest per-query cost at the price of expressiveness. Paradigms~III
and~IV move work into construction and introduce richer retrieval pipelines, with query-time cost below long-context but higher and more variable than flat retrieval. Yet, Fig.~\ref{fig:serving} shows
two orders of magnitude spread in per-query latency across systems on (local) identical
hardware. The remainder of this section reveals comparable spread in construction cost, capability requirements, and storage footprint.

\takeaway{1}{For long-horizon workloads, full-history prefill cost is substantial
and grows with accumulated history. Agent memory systems reduce this cost, but span
two orders of magnitude in per-query serving latency on identical hardware, with
comparable spread in construction costs. These costs are invisible to accuracy metrics, and vary based on the system's construction and
retrieval algorithms.}

\recommendation{1}{Long-horizon agent deployments should treat agent memory system
selection as a system-level decision. Accuracy alone is an insufficient criterion
when systems differ by orders of magnitude in construction cost, serving latency,
and storage footprint.}
\subsection{Construction Dominates the Agent Lifecycle}
\label{sec:hidden-construction-cost}






\begin{figure}[htbp]
    \centering
    \includegraphics[width=\columnwidth]{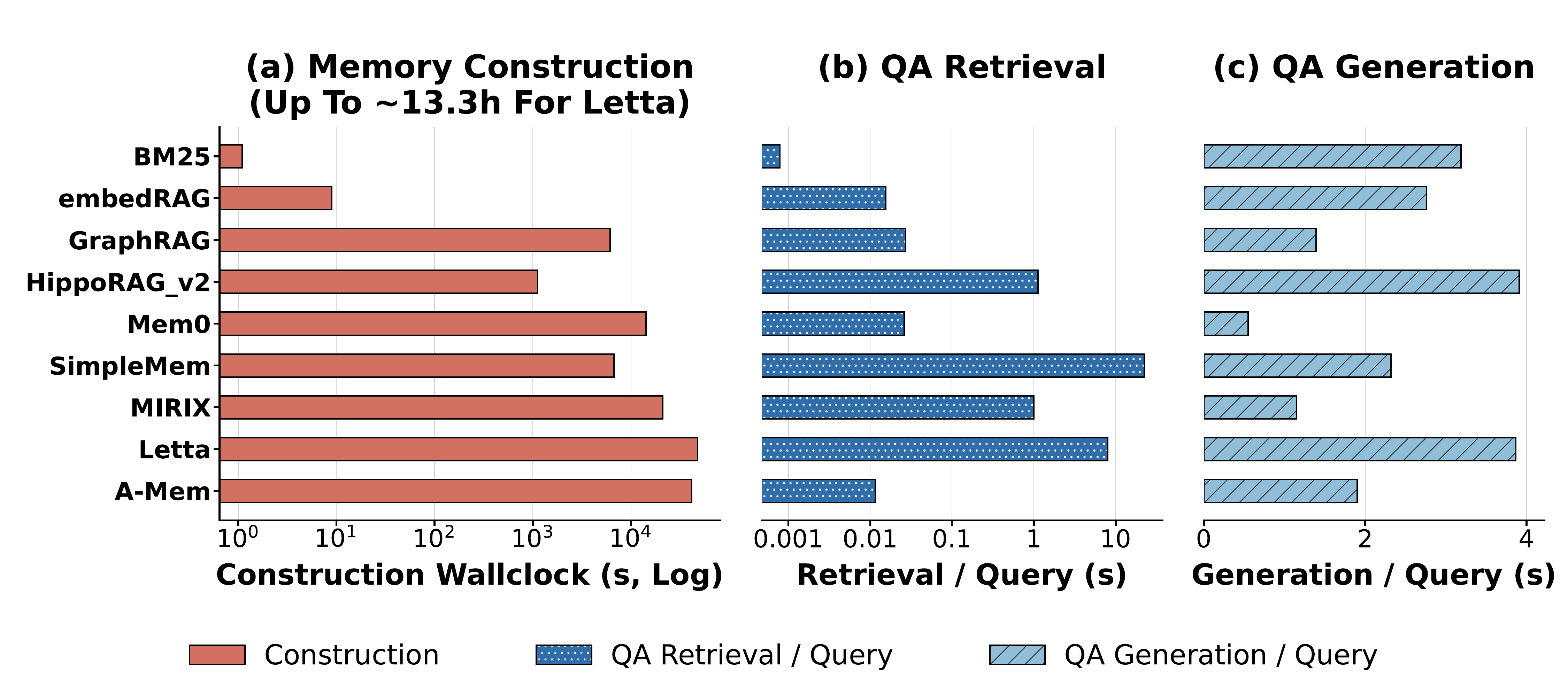}
    \caption{\textbf{Phase cost breakdown.}
    Paradigm~III and~IV agent memory systems shift the majority of end-to-end
    energy into construction, which is invisible to the user at query time.}
    \label{fig:phase-breakdown}
\end{figure}

\begin{table}[htbp]
\centering
\caption{\textbf{End-to-end cost summary on LongMemEval (Qwen3-32B, $n=300$ queries).}
Construct + 300 QA. }
\label{tab:cost-summary}
\resizebox{\columnwidth}{!}{
\begin{tabular}{lrrrrr}
\toprule
\textbf{Agent memory system} & \textbf{Acc.} & \textbf{Wall} & \textbf{Calls} &
\textbf{Total kJ} & \textbf{J/correct} \\
\midrule
BM25        & 47.0 & 16.3m  & 300    & 582    & 4,128   \\
GraphRAG    & 46.0 & 1.83h  & 3,215  & 2,082  & 15,084  \\
HippoRAG v2 & 44.3 & 44.2m  & 2,743  & 1,339  & 10,079  \\
A-Mem       & 42.7 & 11.76h & 19,230 & 14,864 & 116,116 \\
embedRAG    & 39.8 & 14.4m  & 610    & 495    & 4,144   \\
SimpleMem   & 36.0 & 3.92h  & 4,447  & 5,481  & 50,749  \\
Mem0        & 32.0 & 4.02h  & 4,538  & 4,878  & 50,813  \\
Letta       & 27.7 & 14.36h & 18,394 & 15,429 & 185,873 \\
MIRIX       & 20.0 & 6.03h  & 7,655  & 8,678  & 144,629 \\
\bottomrule
\end{tabular}}
\end{table}

\begin{figure}[htbp]
    \centering
    \includegraphics[width=\columnwidth]{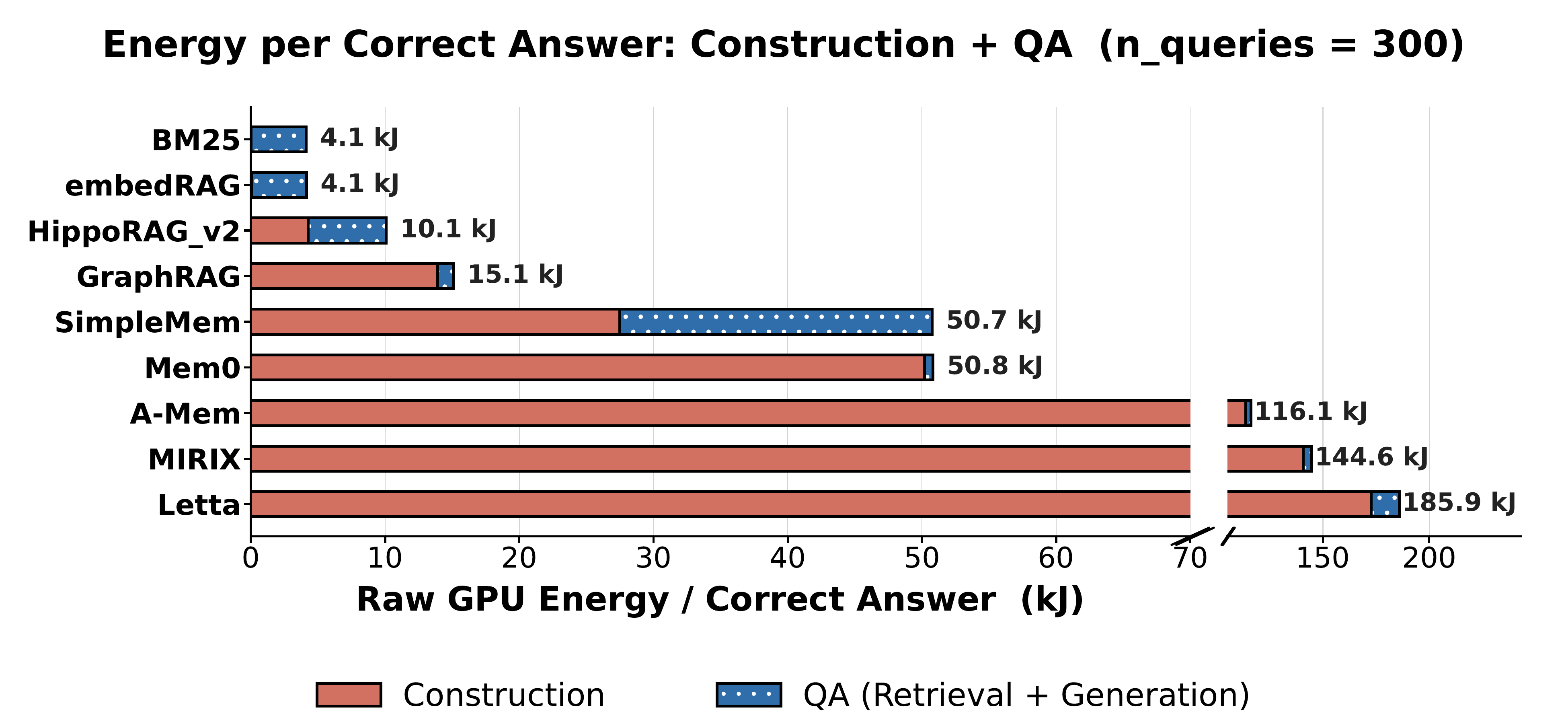}
    \caption{\textbf{Energy per correct answer.}
    Normalizing total energy by correct answers jointly prices construction and
    serving against task quality. The spread across agent memory systems exceeds
    45$\times$.}
    \label{fig:j-per-correct}
\end{figure}

The per-query latency advantage in Fig.~\ref{fig:serving} is conditional on memory having already been constructed. Construction has two cost components that behave differently under scheduling. Energy is invariant: whatever LLM and embedding calls construction makes are paid regardless of when they run. Latency, by contrast, is schedulable. Whether it can be hidden behind asynchronous or background writes depends on the agent memory system's consistency model, a question we address in Sec.~\ref{sec:freshness}. This section quantifies the energy component, which is the unhideable cost the operator pays to deploy each agent memory system.
All experiments presented in this section were measured over five samples of \texttt{LongMemEval\_S\_*}, totaling 1.8 M tokens of history and 300 queries, and run using local vLLM serving (LLM=Qwen3-32B; Embed=Qwen3-Embedding-0.6B).

Fig.~\ref{fig:phase-breakdown} shows construction wall time, as well as the resulting time spent on retrieval and QA. Paradigm~II agent memory systems (BM25, embedRAG) complete construction in under a minute through deterministic indexing. Paradigm~III and~IV systems invoke LLMs repeatedly to extract facts, triples, summaries, or memory updates, pushing wall time from hours (e.g., Mem0 $\sim$4.0h) to over half a day (Letta $\sim$13.3h). The energy consequences follow directly. Table~\ref{tab:cost-summary} shows that end-to-end energy span is over 26.5$\times$, from BM25 at 582\,kJ to Letta at 15,429\,kJ. Isolating construction on one LongMemEval sample widens the gap further. Construction costs 32\,J for
BM25's lexical index, 99\,kJ for HippoRAG~v2's graph build, and 1,050\,kJ for
Mem0's sequential fact-extraction calls, a four-order-of-magnitude spread in
which constructing Mem0's memory for one sample costs 9$\times$ BM25's entire
run on that sample, construction and all 60 queries included.

Fig.~\ref{fig:j-per-correct} normalizes total energy by correct answers, jointly pricing construction and serving against task quality. This is inspired by the Intelligence per Watt metric proposed in \cite{saadfalcon}. BM25
establishes a floor of 4,128\,J per correct answer. A-Mem and Letta reach
116\,kJ and 186\,kJ, a 28--45$\times$ premium that must be justified by
capabilities flat retrieval cannot provide: mutation, conflict
resolution, multi-type routing, or long-range reasoning. \todo{more accurately explain limitations of bm25, bc it seems to perform generally well + TODO: include our numbers from the full 14-dataset sweep when ready.}

\takeaway{2}{Construction energy dominates the agent lifecycle: for most LLM-mediated agent memory systems, construction energy exceeds total query-phase energy across 300 queries. Energy per correct answer exceeds 45 $\times$ across the suite, and agent memory systems with similar accuracy can differ by an order of magnitude on this axis.}

\recommendation{2}{Operators should account for energy across the full agent lifecycle, not just at query time. Construction dominates total energy for most LLM-mediated agent memory systems, which trade construction cost for capabilities such as mutation, structured retrieval, and multi-store routing. Systems with similar accuracy can differ by an order of magnitude in lifecycle energy depending on their construction pipeline.}

\subsection{Construction Is an Overwhelmingly Embedding and Prefill-dominated Workload}
\label{sec:construction-shape}

\begin{figure}[htbp]
    \centering

    \includegraphics[width=\columnwidth]{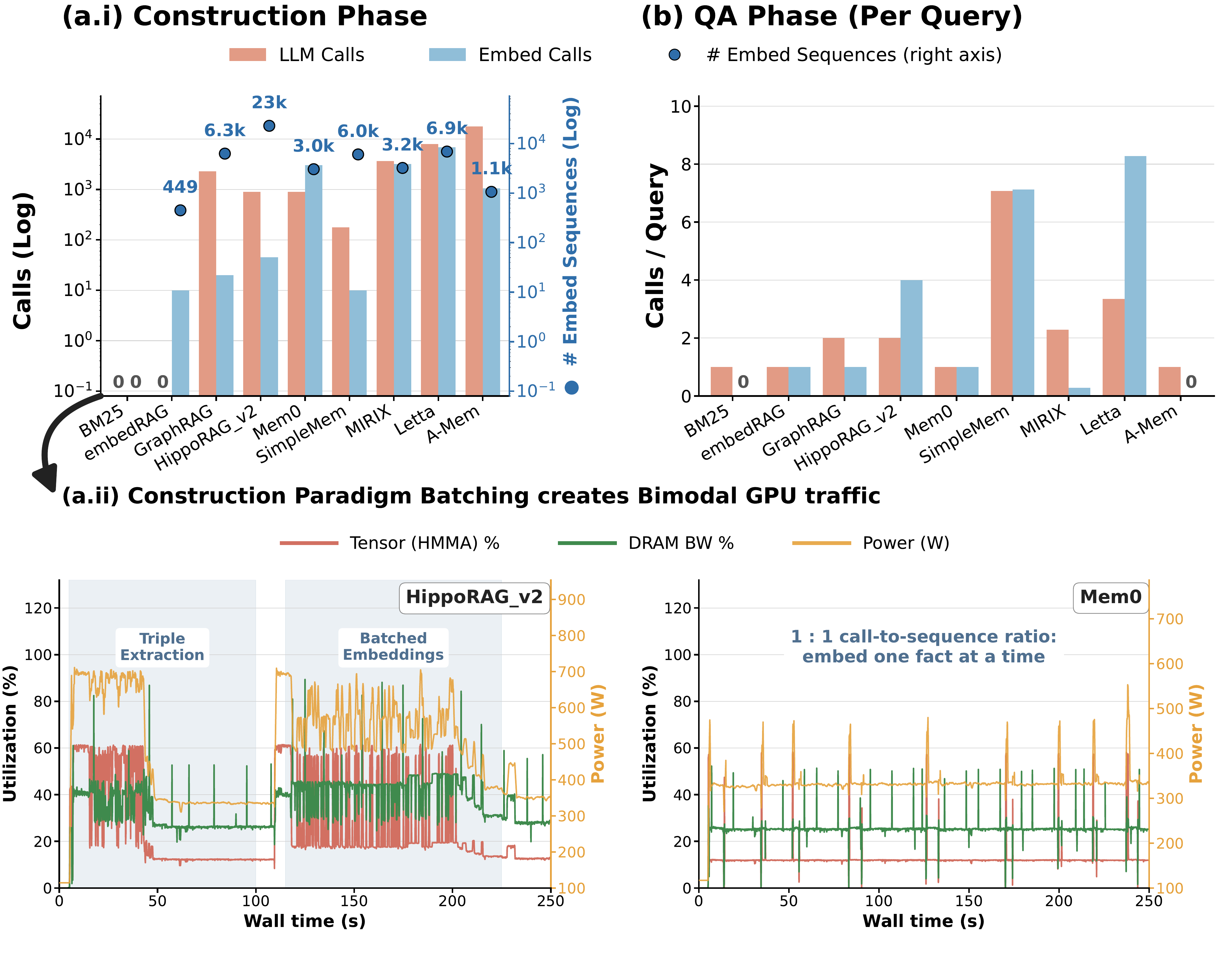}

    \vspace{0.75em}

    \includegraphics[width=\columnwidth]{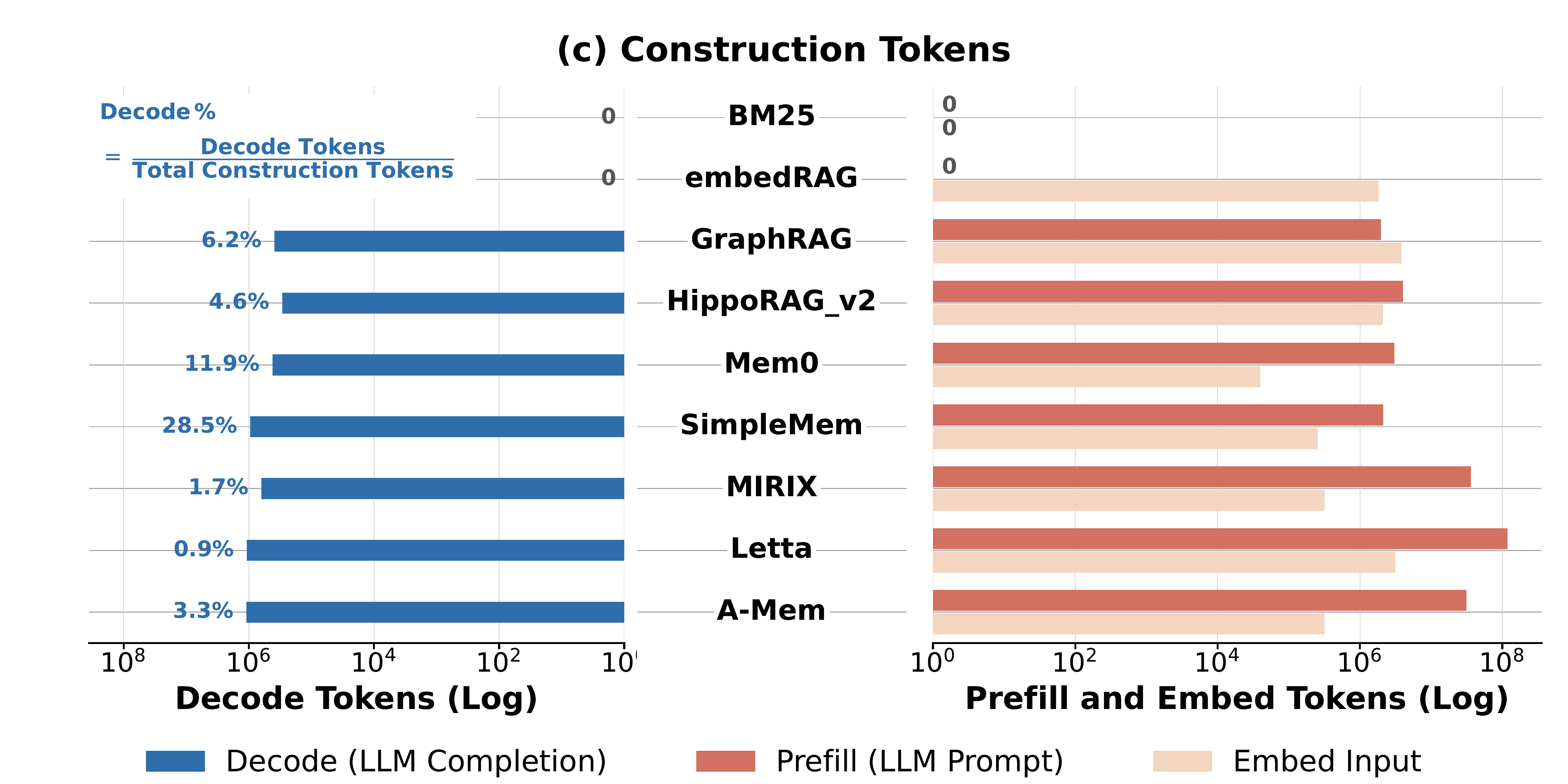}

    \caption{\textbf{Construction call structure and token decomposition.}
    Embedding batching ratios split sharply by taxonomy paradigm: Paradigm~III.a
    generates large-batch offline-indexing traffic; Paradigms~III.b and~IV
    generate sequential per-event traffic on the write-loop critical path.}
    \label{fig:construction-calls-tokens}
\end{figure}

The construction cost identified in Sec.~\ref{sec:hidden-construction-cost} has a
specific computational shape that follows from how each paradigm transforms
interaction history into persistent records.

Fig.~\ref{fig:construction-calls-tokens} decomposes construction traffic into LLM
prompt tokens, completion tokens, and embedding input tokens. Construction is
dominated by embedding and LLM prefill: the embedding model's entire forward pass
is prefill with no decode component, and the LLM construction path reads a window
or chunk and emits a compact structured output (a fact list, a JSON triple set, an
ADD/UPDATE/DELETE decision) whose token count is small relative to the context
consumed. The median decode share of total construction tokens is 4.6\% across
agent memory systems, ranging from 0.9\% (Letta) to 28.5\% (SimpleMem, whose
per-window memory entries are longer). \textbf{Construction is a repeated
long-read, short-write workload.}

This creates a structural conflict with QA serving. QA traffic is
latency-sensitive and optimized around time-to-first-token; construction traffic
is throughput-oriented, processing large input volumes across many independent
chunks with no user-facing latency pressure. Routing both through the same LLM
endpoint means a large construction prefill job occupies KV-cache headroom and
stalls the batch scheduler precisely when a latency-sensitive query arrives.

Embedding traffic is further bimodal by paradigm. Paradigm~III.a agent memory
systems submit large batches to the embedding server: GraphRAG yields $\sim$310
sequences per API call (batching $\sim$10 entity-relation tuples per chunk),
HippoRAG~v2 $\sim$510 sequences per call across three indexed views. Paradigms~III.b
and~IV present the opposite shape: Mem0 must embed each extracted fact before its
similarity search resolves the ADD/UPDATE/DELETE decision, producing a 1:1
call-to-sequence ratio; agentic systems are similarly sequential. An embedding
server configured for batch throughput will create head-of-line blocking for
latency-sensitive write-loop tenants.

The serialized write loops of Paradigms~III.b and~IV
leave much of the device idle: single-tenant, Mem0's construction reaches only
12\% tensor-core activity and 25\% HBM bandwidth, with the GPU mostly idle
between small serialized calls. We test whether this idle capacity can be
recovered by co-scheduling $N$ independent memory-agent tenants against a shared
serving backend, measuring throughput as completed tenant builds divided by time
to final completion. Mem0 scales to 9.6$\times$ at $N{=}10$, with tensor-core
activity rising from 12\% to 19\% and board power from 342\,W to 431\,W. HippoRAG~v2
reaches only 1.4$\times$ at $N{=}10$, with burst peaks unchanged, because its batched construction already ran the backend at its sustained serving
capacity, so additional tenants extend completion time rather than raise
throughput.
Co-tenancy behavior thus splits by paradigm along the same axis as call
structure: consolidating and agentic systems can co-schedule well on shared
backends, while the batched construction of
append-only systems leaves no capacity for concurrent tenants to reclaim.

\takeaway{3}{Agent memory construction is prefill- and embedding-heavy: it
repeatedly reads long chunks or windows and emits compact memory records. Unlike
QA, construction is not on the user-facing decode path, so its cost profile is
closer to background indexing than interactive serving. Embedding traffic splits
further by paradigm, between large-batch offline indexing (Paradigm~III.a) and
sequential per-event writes (Paradigms~III.b and~IV).}

\recommendation{3}{Agent-serving systems should treat memory construction as a
background throughput workload with explicit admission control. Construction jobs
should be rate-limited, batched, or deferred when they would interfere with
latency-sensitive QA. Paradigm~III.a tenants benefit from large embedding batch
sizes. Paradigms~III.b and~IV require per-call latency prioritization on the
embedding path. Because III.b and~IV construction underutilizes the device
single-tenant, operators can co-schedule multiple construction tenants on shared
backends to recover idle capacity near-linearly. III.a construction
already saturates sustained serving capacity with a single tenant and degrades
under contention.}

\recommendation{4}{Construction pipelines should exploit reuse across overlapping
inputs. Windowed and chunked memory systems can reduce repeated prefill cost
through prefix reuse, chunk caching, and batching of independent construction
units.}

The single-GPU configuration
isolates algorithmic behavior from parallelism effects, but the construction
signatures identified above are not artifacts of serving scale. To verify this, we
profiled Mem0 and HippoRAG~v2 under tensor-parallel serving (Qwen3-235B, TP=8,
8$\times$H100), with no modification to the harness. The comparative structure
persists: Mem0's construction remains 13.4$\times$ slower in wall time than
HippoRAG~v2's (32,114\,s vs.\ 2,402\,s), and its tensor-core activity stays below
1\% even at the 95th percentile, because small serialized calls cannot occupy a
sharded deployment any better than they occupy a single device. HippoRAG~v2 retains
its burst signature, reaching 12.4\% tensor-core activity at the 95th percentile.
These signatures are set by construction call structure, not serving scale.

\subsection{Construction-LLM Choice Is Constrained by the Agent Memory System}
\label{sec:construction-llm-choice}

\begin{figure}[t]
    \centering
    \includegraphics[width=\columnwidth]{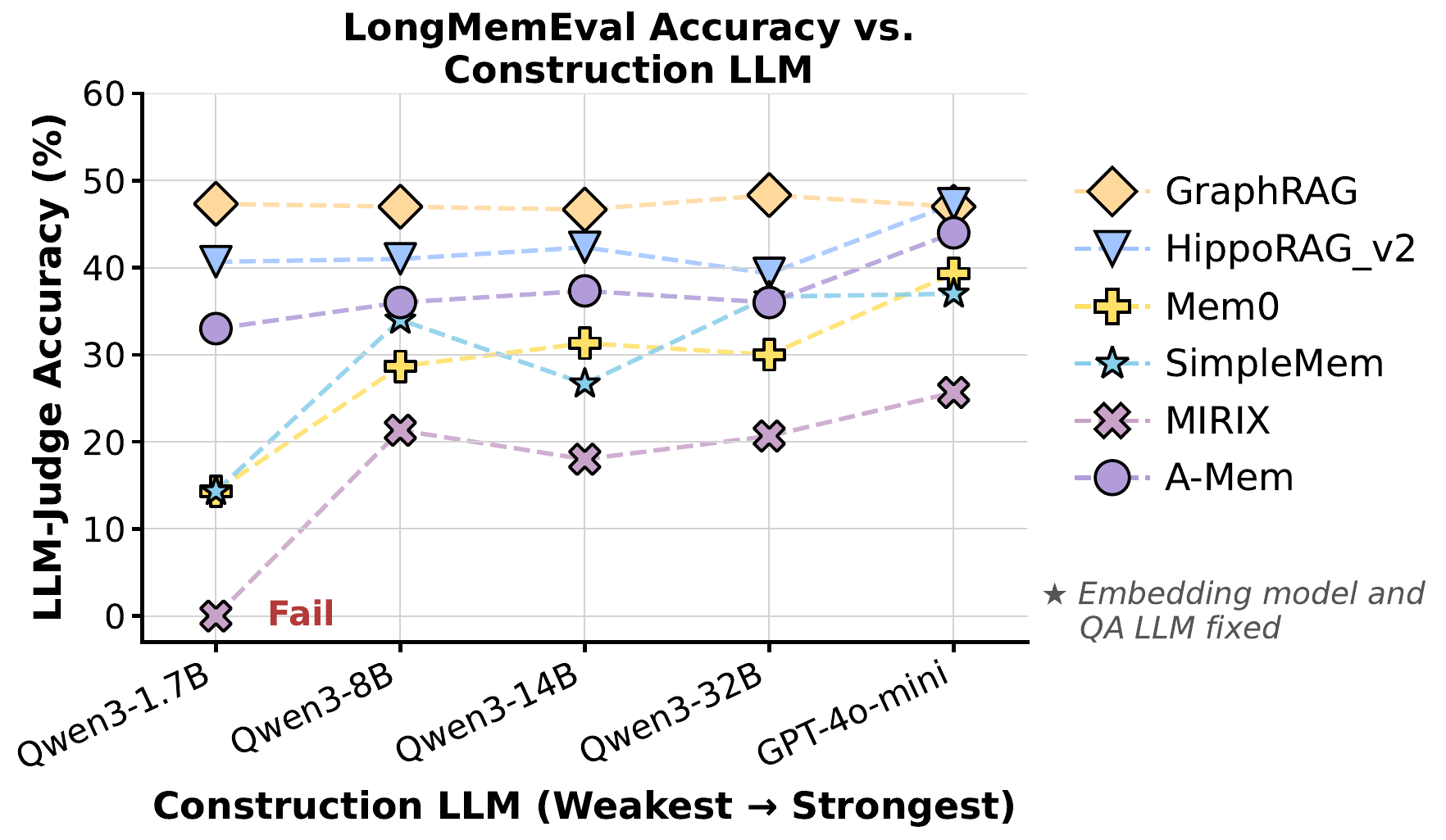}
    \caption{\textbf{Construction-LLM sensitivity.}
    QA LLM is fixed to GPT-4o-mini. Embed is fixed to text-embedding-3-small. Construction LLM is swept for LLM-dependent systems.}
    \label{fig:llm-split}
\end{figure}

Sec.~\ref{sec:hidden-construction-cost} showed that construction energy often dominates the agent lifecycle. A natural cost lever for an operator is to use a smaller, cheaper LLM for construction than for QA. Fig.~\ref{fig:llm-split}
shows that this lever is available for most systems, but its safe range is set by
the construction algorithm's output contracts.

For systems without hard output contracts (Mem0, SimpleMem, A-Mem), accuracy
degrades smoothly as the construction LLM shrinks, permitting continuous
accuracy-cost tradeoffs down to small models. GraphRAG is notably robust,
maintaining $\sim$47--48\% accuracy across the full ladder from Qwen3-1.7B to
GPT-4o-mini, because its entity-relation extraction degrades gracefully rather
than failing structurally. Systems with strict output contracts exhibit a harder floor. MIRIX fails entirely at Qwen3-1.7B, because its pipeline
relies on multiple sub-agent tool calls that require well-formed JSON schemas and
legal tool-call syntax. A model that cannot reliably satisfy these contracts
produces a corrupted store from which the QA model can recover no useful
evidence. 

\takeaway{4}{Construction-LLM downscaling is available for most systems but
its safe range is algorithm-constrained. Systems without hard output contracts
tolerate smaller construction LLMs with
graceful accuracy degradation. Systems with strict output contracts impose a capability floor: falling below it corrupts the
memory store entirely, as seen with MIRIX's complete failure at Qwen3-1.7B.}

\recommendation{5}{Operators should treat the minimum viable construction LLM
as an algorithm-imposed cost floor. For systems with strict output contracts,
this floor must be validated before deployment; falling below it renders the
store unusable. For systems without hard contracts, construction-LLM
downscaling is a cost lever, tunable against an accuracy constraint.}

\subsection{The Construction--Serve--Accuracy Frontier}
\label{sec:pareto}

\begin{figure*}[t]
    \centering
    \includegraphics[width=2\columnwidth]{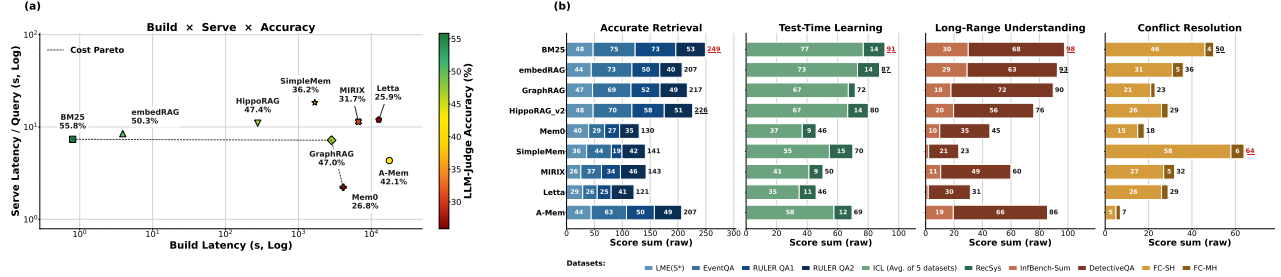}
    \caption{\textbf{(a) Construction--serve--accuracy frontier.}
    Fast construction, fast per-query serving, and high accuracy cannot be jointly
    maximized. Accuracy and cost are macro-averaged over all MemoryAgentBench
    datasets (remote serving, GPT-4o-mini, text-embedding-3-small). (b) Performance of agent memory systems on various task categories in the MemoryAgentBench suite.}
    \label{fig:pareto}
\end{figure*}

Fig.~\ref{fig:pareto} plots construction wall time, per-query latency, and accuracy, each macro-averaged over all MemoryAgentBench datasets. BM25 completes construction in under a second and reaches the highest accuracy in the suite at 55.8\%, but its query time ($\sim$7.4\,s) is behind Mem0, A-Mem, and GraphRAG. The structure-augmented and agentic systems pay substantially more construction cost without overtaking BM25 on accuracy: HippoRAG~v2 reaches 47.4\% at $\sim$277\,s of construction, GraphRAG 47.0\% at $\sim$2{,}850\,s, and A-Mem 42.1\% at $\sim$17{,}666\,s, the most expensive build in the suite. The cost axes also trade against each other. Mem0 attains the lowest per-query latency ($\sim$2.2\,s) by distilling history into atomic facts, but pays $\sim$4{,}108\,s of construction to do so and reaches only 26.8\% accuracy. SimpleMem is the slowest at query time ($\sim$18.4\,s), because its retrieval pipeline adds intent planning, multi-view retrieval, and iterative reflection rounds on top of its compressed entries. No single system is therefore best on all three axes: each occupies a distinct point on the construction--serve--accuracy frontier.

BM25's strong aggregate showing should not be read as a universal result. The macro-average in Fig.~\ref{fig:pareto}a is computed over all datasets in MemoryAgentBench, many of which are recall-heavy and reward the exact-match retrieval that lexical indexing handles well. BM25 has well-documented weaknesses on queries requiring paraphrase matching, multi-hop reasoning, or temporal inference, the task families that motivate Paradigm~III and~IV designs. Fig.~\ref{fig:pareto}b shows this: the aggregate hides sharp per-category variation, and on the harder categories the gap between BM25 and the structure-augmented systems narrows or reverses.

\takeaway{5}{No agent memory system is optimal across construction cost, per-query latency, and accuracy. Agent memory systems with similar accuracy occupy distinct positions on the construction-versus-query cost split.}

\recommendation{6}{Operators selecting an agent memory system should match the construction-versus-query cost split to the workload's query arrival pattern, in addition to matching the agent memory system's capability profile to the dominant task family. High-volume query workloads against stable histories favor agent memory systems that move work into construction. Continuous-ingestion workloads with sparse queries favor agent memory systems with low construction costs.}

\subsection{Inter-Session Construction Creates a Freshness-Latency Tradeoff}
\label{sec:freshness}

\begin{figure*}[t]
    \centering
    \includegraphics[width=2.05\columnwidth]{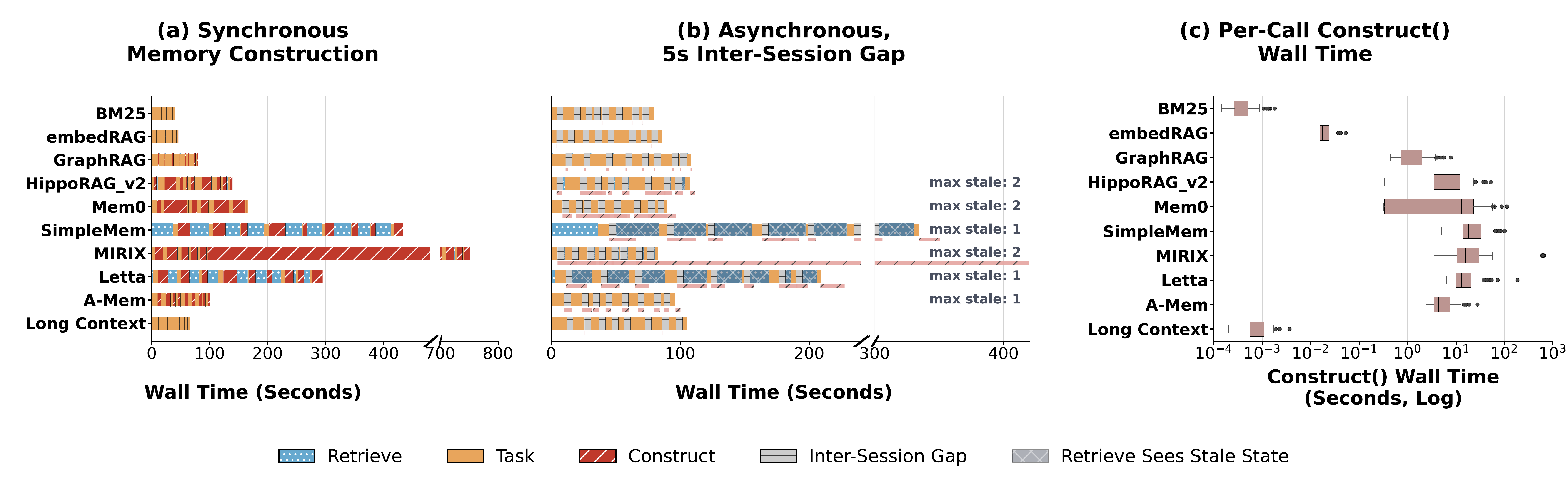}
    \caption{\textbf{Construction scheduling and per-session write latency under
    session arrivals (MemoryArena, physics split, 20 multi-session tasks).}
    Under asynchronous scheduling, slow-construction agent memory systems serve
    queries against stale memory because prior session writes have not yet
    committed. Per-session write latency spans five orders of magnitude, from
    sub-millisecond for BM25 to tens of seconds for Paradigm~IV systems.}
    \label{fig:memoryarena}
\end{figure*}

The preceding sections characterized agent memory in the batch-construct-then-query regime. When sessions instead arrive continuously and later queries depend on memory from earlier ones, the serving system faces a scheduling decision batch evaluation hides.

We characterize this regime using the physics split of MemoryArena~\cite{he2026memoryarena}: 20 multi-session tasks, each a sequence of interdependent subtasks where a later subtask depends on results established by earlier ones. Each subtask is one session, a retrieve-act-write cycle. We capture per-session timing traces and replay them under a controlled 5-second inter-session arrival schedule, emulating synchronous and asynchronous scheduling. All runs use Qwen3-32B-FP8 with Qwen3-Embedding-0.6B.
We define \textit{staleness} as the number of prior sessions not yet persisted to the memory store at query time; a system is \textit{fresh} when staleness is zero. Fig.~\ref{fig:memoryarena}a--b shows one representative task under each mode, one row per system. In panel b, hatched markers indicate queries admitted before the prior write committed. Under synchronous scheduling, slow-construction systems extend the timeline as each session blocks on its own write. Under asynchronous scheduling, SimpleMem, MIRIX, Letta, Mem0, and A-Mem accumulate staleness, retrieving against memory one or more sessions behind the ingestion stream. Paradigm~II systems (BM25, embedRAG) remain fresh in both modes. Fig.~\ref{fig:memoryarena}c shows per-session construction time across all systems,
spanning five orders of magnitude: from $\sim$$10^{-3}$\,s for BM25 to $\sim$10\,s
for Mem0 and MIRIX, with tails exceeding 100\,s for inputs that trigger multiple
consolidation rounds. Construction time alone does not determine feasibility:
retrieval cost during the next session's action loop also contributes, and the
cumulative time must fit within the inter-session interval to avoid a write backlog.
When it does not, the operator faces a forced choice between synchronous scheduling,
which exposes construction latency on the critical path, and asynchronous
scheduling, which admits unbounded staleness.

\takeaway{6}{For agent memory systems where per-session construction time exceeds the inter-session arrival interval, synchronous construction scheduling leads to user-facing latency and memory staleness, thus requiring concurrent (asynchronous) construction scheduling. Per-session construction time alone spans five orders of magnitude
across systems, making it the dominant property for inter-session system selection.}

\recommendation{7}{For inter-session workloads with strict cross-session
dependencies, cumulative construction and retrieval time should be treated as a
hard feasibility constraint. Systems that exceed the arrival interval cannot
satisfy both freshness and latency targets simultaneously. Accuracy alone is an
insufficient selection criterion when this constraint binds.}

\recommendation{8}{Construction cadence should be agent memory system-aware. Append-only
memory can be updated continuously, but consolidating and mutating systems should
monitor marginal per-chunk construction cost and trigger compaction or offline
rebuild when new chunks increasingly spend time comparing against or rewriting
prior records.}

\subsection{Per-User Memory Footprint Growth Diverges Across Agent Memory Systems}
\label{sec:growth}

\begin{figure}[t]
    \centering
    \includegraphics[width=1.\columnwidth]{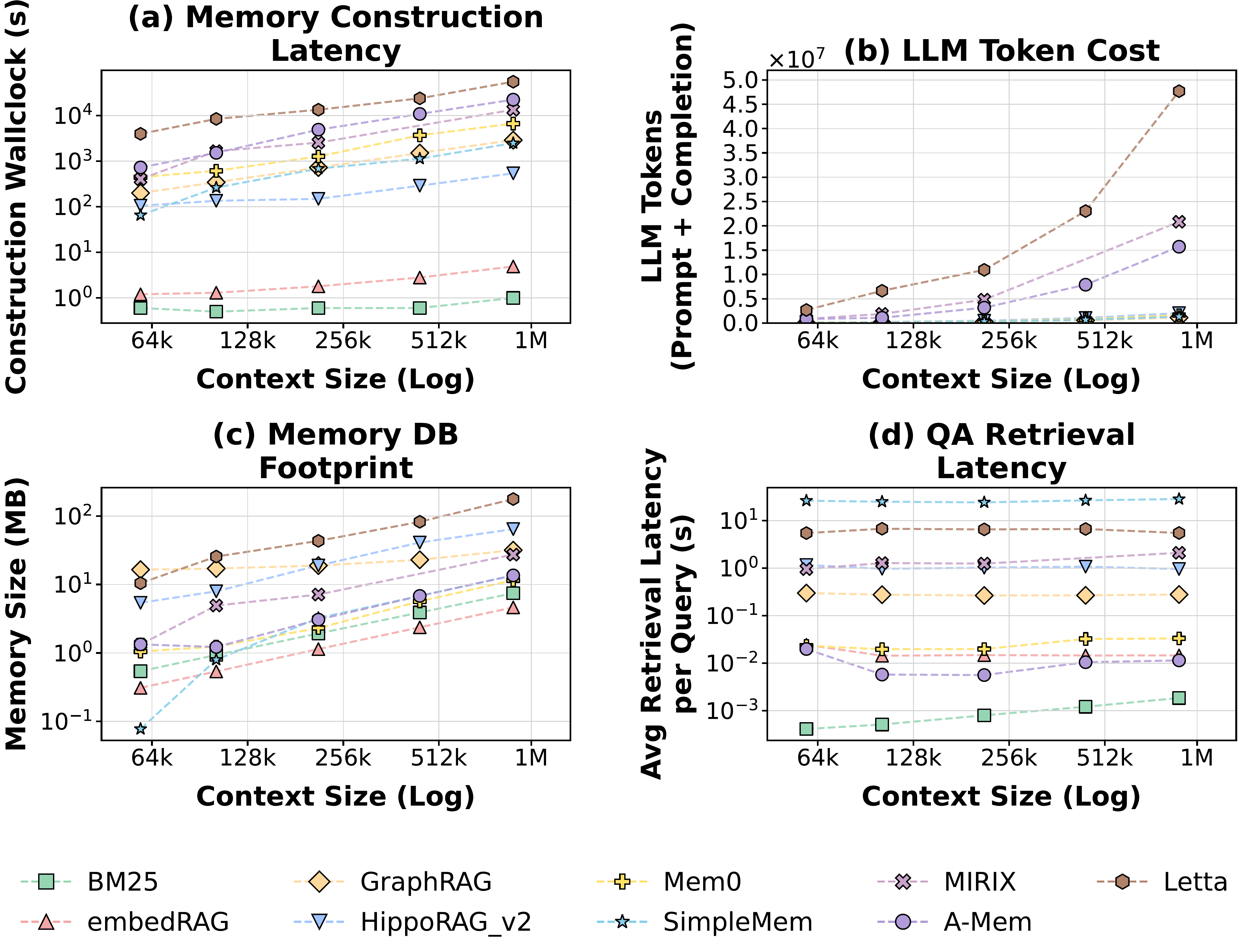}
    \caption{\textbf{Scaling with input length.}
    Construction time, LLM token cost, retrieval latency, and on-disk footprint
    as a single user's history scales from 64\,K to about 1\,M tokens (local serving,
    Qwen3-32B, Qwen3-Embedding-0.6B).}
    \label{fig:scaling}
    \vspace{-0.5cm}
\end{figure}

Agent memory is per-user, unbounded, and accumulates continuously, so per-user
costs compound across users and over deployment lifetime.
Fig.~\ref{fig:scaling} tracks construction time, LLM token cost, retrieval
latency, and on-disk footprint as we scale one user's history from about 64\,K to 1\,M
tokens, using LongMemEval's arbitrary context generation\cite{wu2024longmemeval}.

On-disk footprint (panel c) grows roughly proportionally to content volume for
most systems, with a $\sim$$9\times$ spread at 1\,M tokens. Multi-view designs
inflate the proportionality constant (HippoRAG~v2 reaches $\sim$62\,MB at 1\,M
tokens). Consolidating systems have the potential to compress storage over time
as redundant records are merged, keeping footprint lower than their ingestion
volume would suggest (Mem0 at $\sim$12\,MB). None of the evaluated systems
prune or forget by default, so footprint grows monotonically under default
behavior; bounding fleet storage requires an independent forgetting policy. LLM token cost (panel b) reveals a sharper divergence. Paradigm~IV agentic systems (Letta, A-Mem, MIRIX) scale super-linearly in token
cost: each new ingestion issues tool calls that query and assess the growing
memory store before merging or rewriting records, so per-ingestion token cost
rises with store size and total cost compounds with history length.
Letta in particular diverges steeply beyond 256\,K tokens. Paradigm~II systems
(BM25, embedRAG) incur negligible LLM token cost at all scales. At fleet scale, projecting 1\,M-token footprints to 100\,K users gives a
$\sim$$9\times$ storage spread, from $\sim$0.7\,TB (embedRAG) to $\sim$6.2\,TB
(HippoRAG~v2). But for long-lived deployments the more consequential variable
is the token-cost growth slope: agentic systems whose construction cost
compounds with memory size will become disproportionately expensive as user
histories grow. Retrieval latency (panel d), by contrast, stays nearly flat as the store grows: index-lookup retrieval is sub-linear in store size, decoupling per-query read cost from history length.

\takeaway{7}{Per-user memory footprints vary by up to $9\times$ across systems
at 1\,M tokens, but construction token cost diverges far more sharply.
Paradigm~IV agentic systems scale super-linearly in LLM token cost as growing
memory stores amplify per-ingestion work. Multi-view representations inflate
storage growth; consolidating systems dampen it. An agent's long-term cost is
therefore dictated by its algorithmic growth slope, not its initial footprint.}

\recommendation{9}{Selecting memory for long-lived agents requires evaluating
both baseline footprint and cost growth slope. Agentic systems whose
construction cost compounds with memory size should be paired with active
compaction or summarization policies to prevent unbounded cost escalation.
Because all evaluated systems accumulate state monotonically by default,
operators must add independent pruning or forgetting policies to bound
fleet-scale storage and token costs.}
\subsection{Serving Latency Structure}
\label{sec:retrieval-structure}


\begin{figure}[t]
    \centering
\includegraphics[width=\columnwidth]{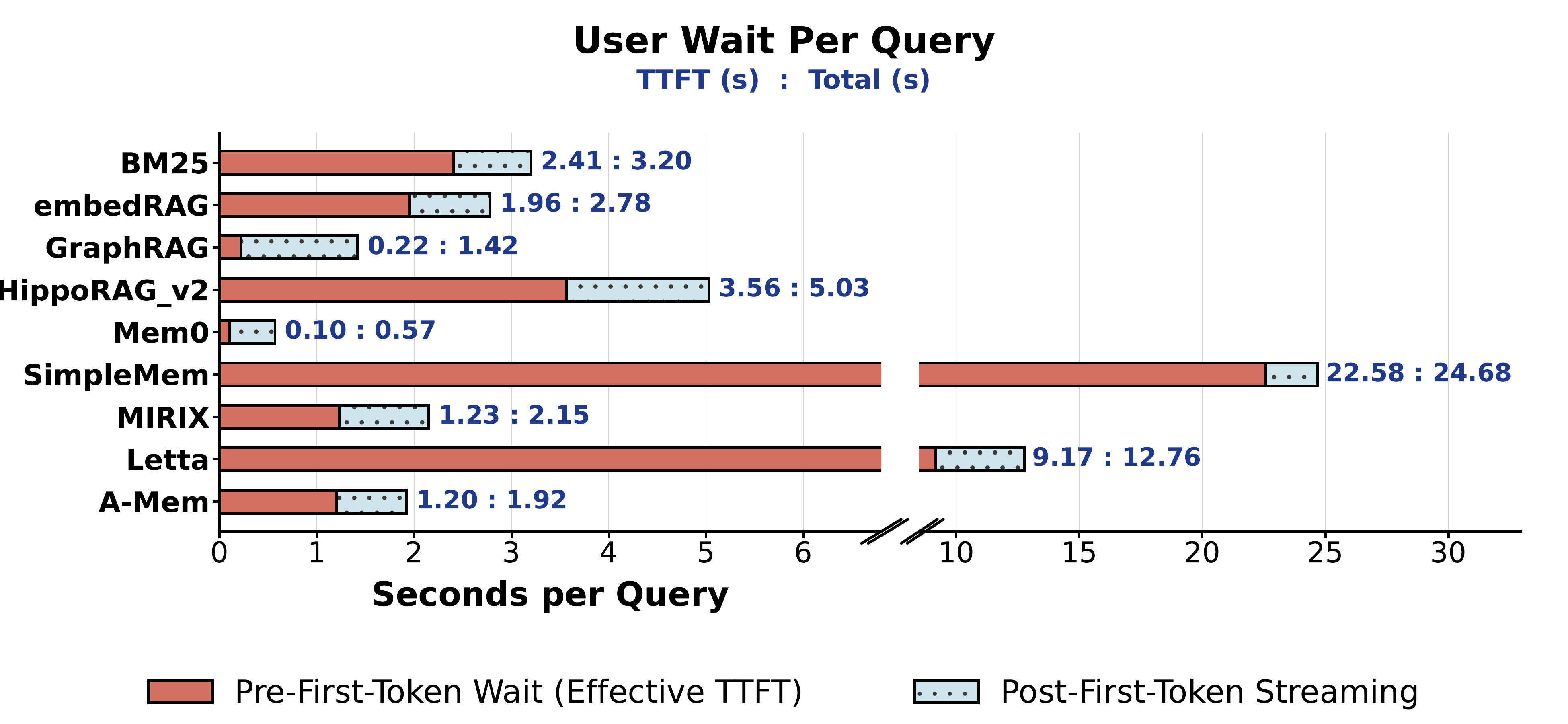}
    \caption{\textbf{Effective time-to-first-token.}
    Pre-answer latency spans two orders of magnitude on identical hardware.
    The dominant variable is retrieval pipeline depth, not LLM serving speed.}
    \label{fig:user-wait}
\end{figure}

\begin{figure}[t]
    \centering
    \includegraphics[width=\columnwidth]{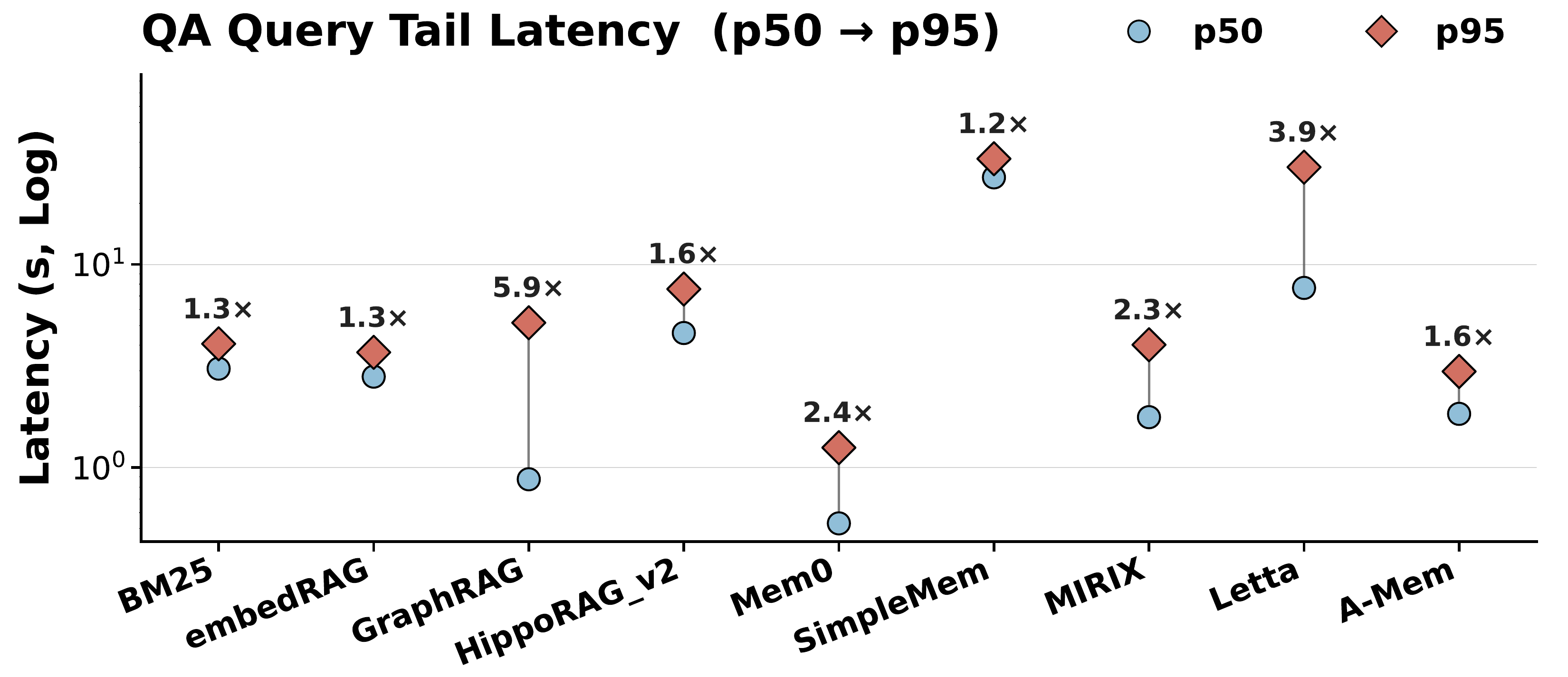}
    \caption{\textbf{QA tail latency by agent memory system.}
    Fixed-depth pipelines (Paradigm~II) have p95/p50 near 1.3$\times$.
    Systems with more complex store structures or query-adaptive pipelines display wider tails at QA time.}
    \label{fig:tail-latency}
\end{figure}

Sec.~\ref{sec:freshness} treated per-session construction and retrieval time as inputs to the scheduling decision. This section explains what sets them. On identical hardware and a fixed corpus, effective time-to-first-token spans two orders of magnitude across systems (Fig.~\ref{fig:user-wait}, $\sim$0.10\,s for Mem0 to $\sim$22.6\,s for SimpleMem), and per-session construction time spans five (Fig.~\ref{fig:memoryarena}c). With the serving stack held constant, the variation comes from how each algorithm bounds its own work. Agent memory systems fall into two regimes, and a system can sit in different regimes for construction versus retrieval. \emph{Algorithm-bounded} phases stop at a point fixed by the algorithm: BM25's single top-$k$ lookup, HippoRAG~v2's fixed passage--entity--fact--PPR sequence, GraphRAG's one extraction call per chunk. Per-phase cost varies with input and graph topology, but its worst case is bounded by static properties the operator can profile in advance. \emph{LLM-bounded} phases continue until the LLM decides they are done: MIRIX's type-specific tool calls, SimpleMem's reflection rounds, Letta's tool-driven memory access, A-Mem's note evolution. When the specification permits arbitrary depth, only an explicit iteration cap bounds cost. The two regimes produce distinct tail behavior. As seen in Fig.~\ref{fig:tail-latency}, deterministic systems have narrow tails: BM25 and embedRAG show p95/p50 ratios near 1.3$\times$, and HippoRAG~v2 only slightly wider at 1.6$\times$ as PPR convergence depends on the query's seed distribution. LLM-bounded systems show the widest tails, reaching 5.9$\times$ for GraphRAG and 3.9$\times$ for Letta, since extra tool calls or reflection rounds drive latency beyond what the corpus alone predicts.

\takeaway{8}{LLM-bounded agent memory systems produce construction and retrieval costs that are less tightly constrained by the input size alone than deterministic paradigms, since runtime LLM decisions can introduce variable numbers of reasoning, tool-use, or refinement steps. External iteration caps or timeouts are thus required to bound worst-case cost.}

\recommendation{10}{Latency-sensitive deployments should treat worst-case latency, on both construction and retrieval, as a selection criterion. Algorithm-bounded systems can be provisioned from worst-case latency measured on representative inputs. LLM-bounded systems require external iteration caps and timeouts, since profiling samples the LLM's behavior on tested inputs but does not bound it on others.}



\section{Discussion \& Conclusion}
This paper presents the first systems characterization of agent memory workloads.
Across ten agent memory systems spanning four paradigms, we find that the dominant
cost in agent memory is not query-time serving but construction, and that
construction is structurally embedding and prefill-dominated, and in direct
tension with latency-sensitive QA traffic when co-located on the same serving
stack. The taxonomy introduced in Sec.~\ref{sec:memory-taxonomy} predicts this
behavior: a system's paradigm determines construction cost shape, embedding traffic
pattern, capability sensitivity to construction-model choice, and retrieval tail
width, making it a systematic lens for deployment decisions beyond the representative
agent memory systems evaluated here.
Several open challenges extend beyond our current scope. Multi-node and multi-agent
deployments introduce consistency and coordination requirements across distributed
memory stores that single-node characterization cannot capture. Multimodal memory, persisting and retrieving images, audio, and structured observations alongside
text, is an emerging and less mature frontier that we expect will amplify the
construction cost, storage footprint, and retrieval complexity challenges
identified here. The taxonomy and profiling methodology developed in this work
provide a foundation for characterizing these richer settings as they mature.

\section*{Acknowledgments}

We thank the staff of the Stanford Marlowe computing cluster and NVIDIA solutions architects Zoe Ryan and Amanda Butler for their continued support. We would also like to thank the Stanford PORTAL and MemoryDAX industrial affiliates program. Finally, we thank the Knight-Hennessy Fellowship and NSF Award 2541050 for funding this research.
\\
\textbf{Statement on Generative AI: }Codebase development was accelerated using Claude Code. Claude was additionally used to assist in improving the presentation of this writing. All technical content, experimental design, analysis,
and conclusions are the work of the authors.


\bibliographystyle{IEEEtranS}
\bibliography{references}

\appendix
\section{Artifact Appendix}

\sloppy

\subsection{Abstract}

The artifact contains a profiling framework for agent memory systems. It
includes the ten systems characterized in the paper behind a common interface.
The framework separates each run into construction and query phases, and each
query into retrieval and generation. Per-call LLM and embedding telemetry is
recorded alongside NVML and GPM counters, enabling phase-level attribution.
The artifact reproduces Figures 2--11 and Table 3, covering latency, accuracy,
wall time, energy, token composition, embedding traffic, hardware utilization,
and storage footprint. Each experiment runs from a single launcher command,
using either a $\sim$6-hour 1-GPU sampling setting or the full published
workload. All measurements were taken on NVIDIA H100 80 GB HBM3 GPUs.

\subsection{Artifact check-list (meta-information)}

{\small
\begin{itemize}
  \item {\bf Algorithm: } Ten agent memory systems across four paradigms,
    long-context, flat RAG, structure-augmented, and agentic control flow.
  \item {\bf Program: } Python profiling framework and Bash launchers.
  \item {\bf Model:} Qwen3-1.7/8/14/32B,
    Qwen3-Embedding-0.6B, GPT-4o-mini, GPT-4.1-mini, text-embedding-3-small.
  \item {\bf Data set: } MemoryAgentBench~\cite{hu2025memoryagentbench}
    (LongMemEval~\cite{wu2024longmemeval}, EventQA, RULER, DetectiveQA,
    InfBench-Sum, MovieRec, FactConsolidation, ICL suite: BANKING77, CLINC150,
    NLU, TREC) and MemoryArena~\cite{he2026memoryarena}.
  \item {\bf Run-time environment: } Ubuntu 22.04, Python 3.12, CUDA 12.8,
    vLLM 0.18.0, NVIDIA driver exposing NVML and GPM counters.
  \item {\bf Hardware: } The harness supports any NVIDIA GPU; reproducing
    reported numbers requires 1$\times$ NVIDIA H100 80 GB HBM3.
  \item {\bf Run-time state: } No other GPU work during evaluation, as device
    power is measured device-wide.
  \item {\bf Execution:} Bash launchers for each sampled pass, see the experimental workflow section.
  \item {\bf Metrics: } Per-phase wall time and energy (J), prefill and decode
    tokens, embedding calls and sequences, tensor-core, SM and HBM activity,
    store footprint (B), time-to-first-token (s), task accuracy.
  \item {\bf How much disk space required (approximately)?: } $\sim$50\,GB for
    the sampling setup and $\sim$85\,GB for the full setup.
  \item {\bf How much time is needed to prepare workflow (approximately)?: }
    $\sim$30 minutes.
  \item {\bf How much time is needed to complete experiments (approximately)?: }
    Sampled experiments take $\sim$6 hours on 1 GPU; OpenAI API runs take
    $\sim$5.5 additional hours, require no GPU, and can run in parallel. The
    full sweep takes $\sim$30 hours on 8 GPUs.
  \item {\bf Publicly available?: } Yes.
  \item {\bf Code licenses (if publicly available)?: } Apache-2.0. Vendored
    systems retain Apache-2.0 and MIT.
  \item {\bf Workflow framework used?: } Bash scripts.
  \item {\bf Archived (DOI): } 10.5281/zenodo.21846239.
\end{itemize}
}

\subsection{Description}

\subsubsection{How to access}

The artifact is available as a GitHub repository: \url{https://github.com/Tambe-Lab/agent-memory-profiling}. It is archived at
10.5281/zenodo.21846239. The repository contains the profiling framework, ten memory
systems, and evaluation scripts. \texttt{REPRODUCE.md} provides detailed
instructions.

\subsubsection{Hardware dependencies}

Most experiments require 1$\times$ NVIDIA H100 80 GB HBM3. Hosted remote-model
experiments need no GPU. The framework also supports 8-GPU tensor parallelism,
though no reported result requires it. Tensor-core, SM, and HBM metrics require
a driver exposing GPM counters.

\subsubsection{Software dependencies}

\texttt{requirements.txt} installs the Python dependencies. Two backends require
additional dependencies: GraphRAG downloads spaCy and NLTK models on first use,
while MIRIX requires PostgreSQL with pgvector on \texttt{PATH}; see
\texttt{REPRODUCE.md}. An OpenAI API key is required for remote-model
experiments and LLM-judge scorers.

\subsubsection{Data sets}

MemoryAgentBench is downloaded from HuggingFace on first use. MemoryArena is a
pinned official submodule patched with our profiling instrumentation. The
LongMemEval variants and 64\,K--1\,M-token histories are included.

\subsubsection{Models}

The Qwen models are served locally by vLLM. \texttt{run/serve.sh} starts a chat
model and an embedding model on 1 GPU. GPT-4o-mini, GPT-4.1-mini, and text-embedding-3-small are accessed through
the OpenAI API.

\subsection{Installation}
\begin{verbatim}
git clone --recurse-submodules <repo-url>
cd agent-memory-profiling
python -m venv .venv
source .venv/bin/activate
pip install -r requirements.txt
echo "OPENAI_API_KEY=sk-..." > .env
\end{verbatim}
The API key is required for remote-model experiments and LLM judging.
\texttt{REPRODUCE.md} provides a $\sim$5-minute smoke test for the environment
and serving stack.

\subsection{Experiment workflow}
\label{sec:workflow}

Experiments use either the full published workload or a sampling setting. The
sampling setting is organized into five passes, A through E:
\texttt{run/reproduce.sh <pass>} runs a pass end to end, including scoring, and
\texttt{run/reproduce.sh -{}-summary <pass>} prints its results keyed by paper
section. Pass A covers the core characterization (\S4.1--4.3, \S4.8) on one
GPU. Pass B runs against hosted OpenAI models and needs no GPU (\S4.5). Pass C
sweeps the construction LLM (\S4.4), Pass D scales history length (\S4.7), and
Pass E runs the inter-session workload in MemoryArena, using the port included
in the artifact (\S4.6).

Passes use local vLLM models, OpenAI models, or both. The primary local mode
colocates Qwen3-32B-FP8 and the embedding model on one GPU. The
construction-LLM sweep holds the QA and embedding models fixed on hosted
endpoints while only the construction LLM varies.

\begin{verbatim}
bash run/serve.sh -m qwen3-32b &
bash run/reproduce.sh A
\end{verbatim}

Results are saved as JSON and profiler traces used to derive phase attribution,
energy, and utilization timelines. \texttt{REPRODUCE.md} gives the commands for
each pass in both settings.

\subsection{Evaluation and expected results}

A sampling run covers a small fraction of the workload, so absolute numbers
will differ from the published ones, but trends should hold. Results should be
compared against the corresponding paper figures. Each result below names the
paper section it supports.

\begin{itemize}
  \item {\bf \S4.1 Serving latency (Fig.~2):} Memory systems answer in a
    fraction of long-context wall time at comparable accuracy, and per-query
    latency spans two orders of magnitude across systems. The sampled setting
    spans over an order of magnitude across four backends, with Mem0 fastest
    and agentic backends slowest. The long-context baseline and full two-order
    spread appear in the full setting.

  \item {\bf \S4.2 Construction dominates the lifecycle (Figs.~3--4,
    Table~3):} For most LLM-mediated systems, construction energy exceeds total
    query-phase energy. The sampled setting shows this for the consolidating and
    agentic backends, while the append-only structure-augmented backend's
    batched construction remains below its query-phase energy.

  \item {\bf \S4.3 Construction call structure (Fig.~5):} Decode accounts for
    a small share of construction tokens, and embedding batch size separates
    the paradigms. The sampled setting shows both: decode remains small relative
    to prefill, while embedding traffic differs by paradigm, with HippoRAG
    batching many sequences per embedding call and Mem0 issuing one call per
    extracted fact at a 1:1 call-to-sequence ratio. GraphRAG's large-batch
    offline-indexing traffic appears in the full setting.

  \item {\bf \S4.4 Construction-LLM sensitivity (Fig.~6):} Systems without
    strict output contracts degrade gracefully as the construction LLM shrinks.
    Across Qwen3-1.7B--32B, GraphRAG maintains similar accuracy, store footprint,
    and construction call count: its entity-relation extraction holds its
    output contract on a far smaller construction model. Systems with strict
    output contracts show a capability floor in the full setting, where MIRIX
    at Qwen3-1.7B produces a store from which the QA model recovers no useful
    evidence.

  \item {\bf \S4.5 Construction--serve--accuracy frontier (Fig.~7):} No system
    leads on construction time, per-query latency, and accuracy simultaneously.
    Across three datasets and four backends, the sampled setting shows no
    backend is Pareto-optimal across all three axes. The full setting extends
    the result across the complete MemoryAgentBench suite.

  \item {\bf \S4.6 Inter-session freshness (Fig.~8):} Per-session construction
    time spans five orders of magnitude, from sub-millisecond for BM25 to tens
    of seconds for Paradigm IV systems. Under asynchronous scheduling,
    slow-construction backends serve stale memory while BM25 remains fresh. The
    sampled run shows both the span and staleness pattern.

  \item {\bf \S4.7 Scaling with input length (Fig.~9):} On-disk footprint
    spreads roughly an order of magnitude across systems at 1\,M tokens,
    Paradigm IV systems scale super-linearly in token cost, and retrieval
    latency stays nearly flat as the store grows. For BM25 and Mem0, footprint
    and token cost grow with history length while retrieval latency stays flat.
    Mem0's LLM token cost grows while BM25's stays near zero.

  \item {\bf \S4.8 Serving latency structure (Fig.~10):} Pre-answer latency
    spans two orders of magnitude, with narrow p95/p50 ratios for
    algorithm-bounded systems and wider ones for LLM-bounded systems. Both hold
    at sampled scale: p95/p50 is near one for algorithm-bounded backends and
    wider for LLM-bounded ones.
\end{itemize}


\subsection{Experiment customization}

The framework supports two additional configurations not used by reported
results. Tensor-parallel serving spans the backbone across 8 GPUs, with the
launcher deriving the parallel size from the requested model. Multi-tenant
co-scheduling places N independent memory agents on one serving backend. For
multi-tenancy, each tenant runs a distinct sample and tenant count is varied.
The hardware profiler samples device counters once, since these cover every
tenant, and attributes host memory to each tenant separately.

A backend and workload are each chosen by a configuration file. Environment
variables enable the hardware profiler and set its sampling rate and output
paths. Model paths and the serving environment are set in one place, so the
launchers run unchanged elsewhere. New memory systems implement a named handler
and inherit instrumentation through patched OpenAI clients.

\subsection{Methodology}

Submission, reviewing and badging methodology:

\begin{itemize}
  \item \url{https://www.acm.org/publications/policies/artifact-review-and-badging-current}
  \item \url{https://cTuning.org/ae}
\end{itemize}

\raggedend
\end{document}